\documentclass[letterpaper,journal]{IEEEtran}
\usepackage{iftex}
\ifPDFTeX
  \usepackage[T1]{fontenc}
  \usepackage[utf8]{inputenc}
\else
  \usepackage{fontspec}
\fi
\usepackage{amsmath,amssymb,amsfonts}
\usepackage{graphicx}
\usepackage{algorithm}
\usepackage{algorithmic}
\usepackage{marvosym}
\usepackage{array,tabularx,multirow,booktabs}
\usepackage{cite}
\usepackage[hyphens]{url}
\usepackage{textcomp}
\usepackage{stfloats}
\usepackage[hidelinks]{hyperref}
\newcommand{\tabresult}[3]{%
  \makebox[1.8em][r]{#1}%
  \makebox[2.7em][l]{%
    \textsubscript{\scriptsize $\pm$#2}%
    \textsuperscript{#3}%
  }%
}
\newcolumntype{Z}{>{\centering\arraybackslash}X}
\ifPDFTeX
  \DeclareUnicodeCharacter{00D7}{\ensuremath{\times}}
  \DeclareUnicodeCharacter{00B0}{\textdegree}
  \DeclareUnicodeCharacter{00B1}{\ensuremath{\pm}}
  \DeclareUnicodeCharacter{2212}{\ensuremath{-}}
\fi
\begin{document}
\title{MCDA: Multi-View Graph Contrastive Learning with Dynamic Topology Augmentation for Functional MRI Analysis}
\author{Hao~Zhang,~\IEEEmembership{Graduate Student Member,~IEEE}, Xiaoyun~Liu, Shuo~Huang, Yue~Ma,\\
Liang~Cao\textsuperscript{\Letter}, Mingxia~Liu\textsuperscript{\Letter},~\IEEEmembership{Senior Member,~IEEE}, and Li~Zhang\textsuperscript{\Letter}%
\thanks{Hao Zhang is with the College of Biomedical Engineering, Fudan University, Shanghai 200433, China, and also with the College of Information Science and Technology \& Artificial Intelligence, Nanjing Forestry University, Nanjing 210037, China.}%
\thanks{Shuo Huang and Li Zhang are with the College of Information Science and Technology \& Artificial Intelligence, Nanjing Forestry University, Nanjing 210037, China.}%
\thanks{Xiaoyun Liu is with the Department of Psychosomatics and Psychiatry, Zhongda Hospital, School of Medicine, Jiangsu Provincial Key Laboratory of Brain Science and Medicine, Southeast University, Nanjing 210009, China.}%
\thanks{Yue Ma is with the College of Clinical Medicine, Jiangsu Health Vocational College, Nanjing 210018, China.}%
\thanks{Liang Cao is with Taian Tumor Hospital, Taian 271000, China.}%
\thanks{Mingxia Liu is with the Department of Radiology and Biomedical Research Imaging Center (BRIC), University of North Carolina at Chapel Hill, NC 27599, USA.}%
\thanks{\begingroup\raggedright
\textsuperscript{\Letter}\,\textit{Corresponding authors:} Liang~Cao, Mingxia~Liu, and Li~Zhang.
E-mail: \href{mailto:caoliang198009@126.com}{caoliang198009@126.com}; \href{mailto:mingxia_liu@med.unc.edu}{mingxia\_liu@med.unc.edu}; \href{mailto:lizhang@njfu.edu.cn}{lizhang@njfu.edu.cn}.\par
\endgroup}%
}

\maketitle

\begin{abstract}
Resting-state fMRI (rs-fMRI) is a vital non-invasive tool for diagnosing brain disorders and uncovering pathological mechanisms. 
However, existing graph contrastive learning (GCL) methods often underexploit augmentation histories across training epochs and the relationships among different augmented views.
In this paper, we introduce a topology-guided Multi-view graph Contrastive learning framework with Dynamic graph topology Augmentation (MCDA) for fMRI analysis. 
Unlike traditional GCL that uses fixed noise or edge-dropping, MCDA employs a Dynamic Augmentation Update (DAU) module that iteratively refines the augmented topology. 
Specifically, we incorporate self-aware and cross-sample augmentation strategies to simultaneously capture individual-specific functional signatures and inter-subject heterogeneity. 
We then design a topology-guided multi-view graph contrastive learning scheme that explores interrelationships not only between the original network and its augmented views, but also across different augmented views to maximize structural information preservation. 
We characterize the score-level optimization directions and establish a lower bound for the contrastive objective. The framework combines classification and contrastive losses for brain disorder diagnosis and examines discriminative functional connectivity patterns as potential biomarkers. 
Experiments on two public rs-fMRI datasets comprising 1,236 subjects demonstrate that MCDA outperforms the state-of-the-art methods. The code has been made publicly available at: \url{https://github.com/goodcodingagoodboy/MCDA}.
\end{abstract}


\section{Introduction}\label{sec:introduction}

\IEEEPARstart{R}{esting-state} functional MRI (rs-fMRI) is a powerful non-invasive tool for uncovering the pathological mechanisms of brain disorders~\cite{liu2022resting,ma2026multi}. 
Rs-fMRI data can be modeled as non-Euclidean graphs that capture functional connectivity patterns and interactions between brain regions of interests (ROIs), facilitating automated diagnosis of brain disorders like major depressive disorder (MDD) and autism spectrum disorder (ASD)~\cite{marx2023major}. 
% However, the practical deployment of these graph-based models is often constrained by a significant \emph{data scarcity} challenge: expert annotations are costly, and patient privacy concerns (along with the rarity of certain diseases) limit the availability of large-scale labeled datasets. 

Various graph-based learning methods have been developed to extract discriminative representations from fMRI data, such as BrainGNN~\cite{li2021braingnn} and graph transformer-based foundation model~\cite{wang2026graph}. 
Li \textit{et al.}~\cite{li2021braingnn} proposed a clustering-based embedding method in the graph convolutional layer and provided users the flexibility to interpret different levels of biomarkers through graph node pooling and several innovative loss terms to regulate the pooling operations.
Instead of treating brain networks as graphs for processing, some studies also convert rs-fMRI from graph structures into low-dimensional vectors for embedded learning. 
Wang \textit{et al.}~\cite{wang2026graph} proposed a graph transformer-based foundation model for brain functional connectivity networks (FCN) that integrates node and edge embeddings to extract versatile features.
% While effective, these supervised or generative models typically require substantial annotated samples to achieve robust generalization, which is often unattainable in clinical practice. 

Graph contrastive learning (GCL) has been investigated for fMRI representation learning and brain disorder analysis~\cite{wang2023unsupervised,zong2024new}.
By treating each brain network as a graph structure, GCL typically generates different augmented views of a brain network and then optimizes feature space via a contrastive loss, thereby boosting robustness and accuracy in downstream tasks. 
Beyond generating augmented views at the current training step, we investigate how their evolution across epochs can support history-guided augmentation refinement.
Furthermore, different augmented views are expected to express distinct functional connectivities from brain networks, enabling a comprehensive representation of topological structures. 
Nevertheless, most existing GCL methods~\cite{wang2023unsupervised,tian2024multi,zhang2025multi} fail to model the interrelationships among augmented views and tend to overlook feature expression between augmented networks.
Those limitations often affect discriminative power of graph representation, 
ultimately affecting the accuracy and robustness of task-specific predictions.

\begin{figure*}[!t]
    \centering
    \includegraphics[width=\textwidth]{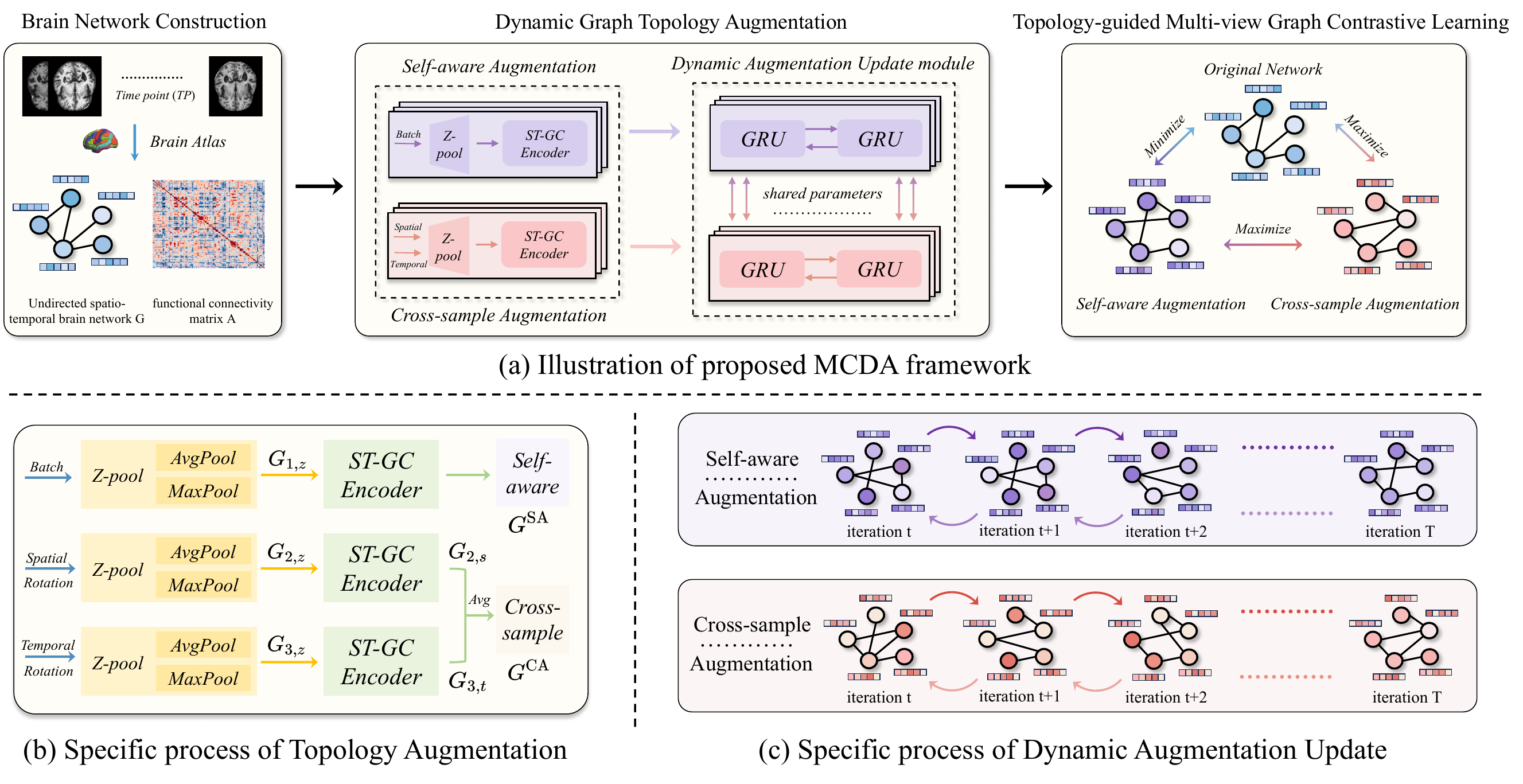}
    \vspace{-1.5em}
    \caption{Overview of the proposed MCDA framework for rs-fMRI analysis. (a) The framework comprises brain network construction, dynamic graph augmentation, and topology-guided multi-view graph contrastive learning. (b) Self-aware augmentation provides subject-specific modulation, while cross-sample augmentation introduces spatial and temporal perturbations from other subjects. (c) The DAU module uses a shared BiGRU to summarize cross-epoch augmentation histories and sigmoid-gated fusion to refine current augmented views using branch-specific historical references.}
    \vspace{-1em}
    \label{fig:framework}
\end{figure*}

To address these challenges, we introduce a topology-guided Multi-view graph Contrastive learning framework with Dynamic graph topology Augmentation (\textbf{MCDA}) for brain disorder identification and fMRI analysis. 
As illustrated in Fig.~\ref{fig:framework}(a), the MCDA framework employs dynamic graph topology augmentation to continuously generate two complementary brain networks, integrating self-aware and cross-sample augmentation to model both subject-specific traits and functional variability across individuals. 
Then, we propose a topology-guided multi-view graph contrastive learning scheme, which can capture not only the interrelationships between each augmented view and the original network but also the interrelationships between different augmented views. 
A similarity-score gradient analysis characterizes the optimization directions of these constraints, and a boundedness analysis establishes a lower bound for the contrastive objective.
We evaluate the MCDA on two public rs-fMRI datasets (1,236 subjects), with experimental results validating that our MCDA surpasses several state-of-the-art (SOTA) approaches in automated brain disorder identification.

The main contributions of this work are summarized below:
\begin{enumerate}
    \item We develop a dynamic augmentation update module that adaptively refines augmented brain networks through history-guided gated fusion.

    \item The proposed dual-augmentation strategy combines subject-specific modulation with cross-sample spatial and temporal perturbations to generate complementary brain network views.

     \item A topology-guided multi-view contrastive scheme balances consistency and diversity through constraints between original views and across augmented views.

    \item We characterize the optimization directions of the contrastive constraints through similarity-score gradient analysis. Experiments on two public rs-fMRI datasets of 1,236 subjects assess classification performance and explore model-derived brain region associations.

\end{enumerate}

The remainder of this paper is organized as follows. We briefly review the most relevant work in Section~\ref{sec:related work}. Section~\ref{sec:methodology} interprets our proposed MCDA framework, while Section~\ref{sec:experiment} covers materials, experimental settings and experimental results. Section~\ref{sec:discussion} examines key components, parameter sensitivity, brain region associations, and limitations of MCDA. Finally, this paper is concluded in Section~\ref{sec:conclusion}.

\section{Related Work}\label{sec:related work}

\subsection{Graph Augmentation of FMRI-Derived Brain Networks}

Graph augmentation generates diverse brain network views to support robust representation learning from rs-fMRI data.
Existing techniques can largely be categorized as static epoch-based augmentation strategies. 
Early representative methods~\cite{wang2023unsupervised, oh2020time} perform graph augmentation directly on the original BOLD time series using window slicing to generate diverse training samples while preserving temporal continuity. 
More recently, deep generative models have been employed to produce diverse views that conform to the data distribution. 
For instance, Wang \textit{et al.}~\cite{wang2025self} integrated diffusion models with graph augmentation, simultaneously modifying both the topological structure and node features to enhance brain network representations. 
% Their Graph Diffusion Augmentation (GDA) module consisted of a noise unit and a denoising neural network. The denoising network predicted the original graph, and a posterior distribution and noise unit were used to sample a new graph based on the prediction to generate augmented brain networks.
Li \textit{et al.}~\cite{li2025decoding} perform graph augmentation via multi-view contrastive self-supervised learning, where a learner-instructor framework enforces consistency and enables self-updating to mine hidden topological connections from raw fMRI data.

Augmentation strategies driven solely by current representations do not explicitly exploit the evolution of augmented views across training epochs. In contrast, MCDA summarizes recent augmentation histories using a shared BiGRU and adaptively fuses current views with historical references through sigmoid gating, enabling history-guided refinement beyond current-step view generation.

%However, by focusing solely on the current training epoch, these methods neglect the effective information propagation and preservation of discriminative feature representations during the augmentation process. Given this fact, we propose dynamic graph topology augmentation to capture richer structural information in the augmented views. 
  
% However, current graph augmentation techniques for brain networks mainly focus on rs-fMRI temporal slices and static topology enhancement using generative deep networks, often neglecting subject-specific and inter-subject functional connectivity variations and dynamic information propagation during training. Given this fact, we propose dynamic self-aware and cross-sample topology augmentation to generate augmented brain networks and integrate it with contrastive learning, to collectively advance brain  disease recognition tasks.

\subsection{Graph Contrastive Learning for fMRI analysis}

Graph contrastive learning (GCL) analyzes relationships between augmented views to learn brain network representations for rs-fMRI-based brain disorder diagnosis.
By maximizing the agreement between differently augmented views of the same brain network, GCL encourages the model to capture invariant and robust functional patterns~\cite{zhang2025graph}.  
Early works in this domain primarily utilized single-view settings, focusing on temporal or functional perturbations to project brain networks into a discriminative feature space~\cite{xiong2024self}. 
More recently, research has shifted toward multi-view contrastive learning to leverage diverse perspectives of brain connectivity. 
For instance, Li \textit{et al.}~\cite{li2025decoding} developed a learner--instructor framework for unsupervised brain topology inference, using multi-view contrastive self-supervision to adaptively refine ROI interconnections. 
Zong \textit{et al.}~\cite{zong2024new} proposed an alternative multi-view approach that computes similarity metrics at both node and graph level perspectives between the original brain networks and their graph diffusion variants, enabling simultaneous capture of node and edge level features. 
% ~\cite{peng2022gate} proposed a method that embeds brain networks using a GCN encoder after performing data augmentation. 
% To avoid feature collapse in the embedding space, they introduced an input consistency regularization term, ensuring that the low-dimensional features can preserve their representation capability. 

These studies illustrate the value of multi-view inputs and multi-level representations in GCL. Nevertheless, constructing multiple views does not by itself specify how consistency and diversity should be balanced across augmented representations. Building on our preliminary multi-view formulation~\cite{zhang2025multi}, MCDA jointly regulates original to augmented and augmented to augmented relationships. Rather than uniformly maximizing agreement, it combines self-aware to original consistency with relative cross-sample to original and two augmented views diversity, encouraging complementary representations from the two augmentation branches.

%While multi-view graph contrastive learning facilitates the extraction of discriminative features across diverse perspectives, many approaches fail to capture interrelationships among augmented views and original view, leading to the loss of a comprehensive representation covering both topology structures and dynamic connectivity. This work fills this gap by explicitly considering the comprehensive interrelationships between augmented views and minimizing their cross-view similarity, thereby achieving effective separation of discriminative features and maximal preservation of structural information across views.

\section{Methodology}\label{sec:methodology}

In this work, we propose a topology-guided multi-view graph contrastive learning framework with dynamic graph topology augmentation (MCDA) for rs-fMRI-based brain disorder analysis. The framework is composed of three key components: 1) brain network construction, 2) dynamic graph topology augmentation, and 3) topology-guided multi-view graph contrastive learning. An overview of the proposed MCDA framework is presented in Fig.~\ref{fig:framework}(a). More details are elaborated below.

\subsection{Brain Network Construction}\label{sec:brain network construction}

To characterize brain dynamics and functional interactions, we use rs-fMRI to construct an undirected spatio-temporal brain network \(\mathbf{G} \) and a functional connectivity matrix \( \mathbf{A}\)~\cite{wang2021modeling}. For each subject, blood-oxygen-level-dependent (BOLD) signals from rs-fMRI represent an undirected spatio-temporal network \( \mathbf{G} = (\mathbf{V}, \mathbf{E}) \in \mathbf{R}^{ T\times N} \). The vertex set \( \mathbf{V} = \{ v_{ti} \mid t = 1, \dots T;  i = 1, \dots, N \} \) includes each ROI (region of interest) at every time point \( t \), where  \( T \) denotes the number of time points during scanning, and \( N \) is the number of brain regions. 

The edge set \( \mathbf{E} \) contains two subsets from both temporal and spatial connectivity. For each subject, the functional connectivity matrix $\mathbf{A}\in\mathbb{R}^{N\times N}$ is constructed by computing Pearson correlations between the mean BOLD time series of all ROI pairs.

\subsection{Dynamic Graph Topology Augmentation} \label{sec:dynamic topology augmentation}

To fully leverage the topology structure of the brain network and optimize the propagation of discriminative information in augmented views, we propose a dynamic self-aware and cross-sample topology augmentation to generate two complementary augmentations of brain network for both individual-specific characteristics and inter-subject functional heterogeneity.

\subsubsection{Self-aware and Cross-sample Topology Augmentation}

\begin{figure}[!t]
    \centering
    \includegraphics[width=\columnwidth]{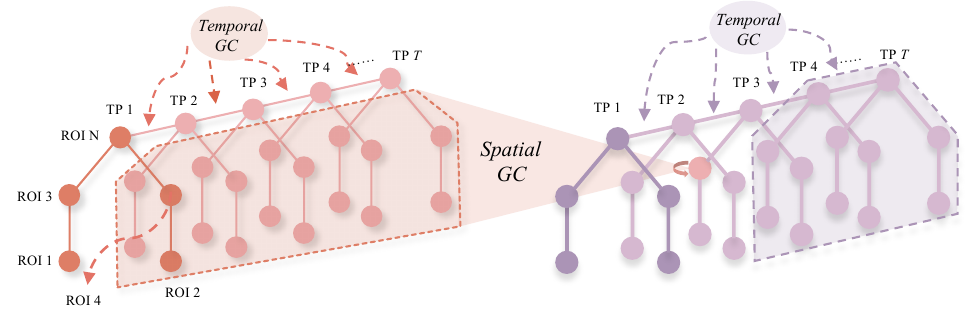}
    \vspace{-1em}
    \caption{The spatio-temporal GC Encoder consists of spatial and temporal graph convolution parts. Here, TP refers to time points, and ROI represents brain region nodes.}
    \label{fig:STGC Encoder}
    \vspace{-1.5em}
\end{figure}

Considering the effect of sample sizes in topology augmentation, we extend the brain network \( \mathbf{G} \)  and \( \mathbf{A}\) into three dimensions, represented as \( \mathbf{R}^{B \times T \times N} \) and \( \mathbf{R}^{B \times N \times N} \). Here, B refers to the batch size used during each training epoch. And, the detailed procedure of self-aware and cross-sample topology augmentation is presented in Fig.~\ref{fig:framework}(b).

As for self-aware augmentation, first, the input \(\mathbf{G}\) is fed into Z-pool layer~\cite{misra2021rotate}, producing the \(\mathbf{G}_{z}\). The transformation equation for the Z-pool layer is given by:
\begin{equation}
\begin{aligned}
\mathbf{G}_{z}
&=\operatorname{Z\text{-}pool}(\mathbf{G})=\operatorname{MaxPool}(\mathbf{G})\oplus\operatorname{AvgPool}(\mathbf{G})\\
&=\left[\max_{t,n\in R}(\mathbf{G}_{t,n})\oplus\frac{1}{|R|}\sum_{t,n\in R}\mathbf{G}_{t,n}\right],
\end{aligned}
\label{eq:zpool}
\end{equation}
where \( \mathbf{G}_{t,n} \) denotes the BOLD signal at time $t$ and ROI $n$ in brain network, and \( R \) represents the local neighborhood in the spatio-temporal domain, with size \( |R| \). MaxPool extracts the maximum response within each local spatio-temporal neighborhood, while AvgPool computes the mean response over the same region. The operator $\oplus$ combines these complementary local statistics to form the pooled representation. 

Next, the Spatio-temporal Graph Convolutional (ST-GC) encoder, as illustrated in Fig.~\ref{fig:STGC Encoder}, is employed to encode \(\mathbf{G}_{z}\). It consists of two components---spatial and temporal parts---which jointly capture the structural and dynamic characteristics of brain functional connectivity. We use the \( \mathbf{G}_{z}\) as the initial input of the first spatial layer \(\mathbf{G}_{s,in}^{1}\). The output feature of the \(l\)-th spatial GC encoder is expressed as:
\begin{equation}
\mathbf{G}_{s,out}^{l} = \Lambda^{-\frac{1}{2}} (\mathbf{A}+\mathbf{I}) \Lambda^{-\frac{1}{2}}\mathbf{G}_{s,in}^{l} W_{S}^{l},
    \label{eq:spatial_gc_unit}
\end{equation}
For clarity, the subject and time indices are omitted in Eq.~(2). Here, $\mathbf{G}_{s,\mathrm{in}}^{l}\in\mathbf{R}^{N\times C_l}$ contains the features of all ROIs for one subject at one time point, where $C_l$ is the input feature dimension. $\mathbf{I}$ is the identity matrix, and $\boldsymbol{\Lambda}$ is diagonal with
$\Lambda_{ii}=\sum_{j=1}^{N}(A_{ij}+I_{ij})$.
The normalized adjacency acts on the ROI dimension, while $\mathbf{W}_{S}^{l}$ transforms the feature channels. When it comes to the temporal GC encoder, the spatial features from the same layer \( \mathbf{G}_{s,out}^{l} \) serve as inputs to a temporal convolution operation, and the entire process can be formulated as follows:
\begin{equation}
    \mathbf{G}_{t,out}^{l} = \psi \left( \phi \left( \mathbf{G}_{s,out}^{l} \right) * W_{T}^{l} \right).
    \label{eq:temporal_gc_unit}
\end{equation}
where \( * \) denotes the temporal convolution operation, \( W_{T} \) represents the convolution operation for each ROI with convolution kernel of \( k\times\)1. Here, \(k=\text{11}\) represents the temporal window of connected neighboring time steps. \(\mathbf{G}_{t,out}^{l}\) denotes the output of the \(l\)-th layer of the temporal part.  \(\phi\) denotes BN-ReLU operations. \(\psi\) represents BN-Dropout operations, which use a dropout rate of 0.5 to randomly discard features, preventing overfitting during training. Let $\Phi^{(0)}(\cdot,A)$ denote the ST-GC encoder for self-aware augmentation, as described in Eq.~\ref{eq:spatial_gc_unit} and~\ref{eq:temporal_gc_unit}.
Following the multiplicative modulation in~\cite{zhang2025graph}, the self-aware view is obtained as
\begin{equation*}
\mathbf{G}^{\text{SA}} = \mathbf{G}_{\text{z}} \odot \sigma\!\left(\Phi^{(0)}(G_z,A)\right),
\end{equation*}
where $\sigma$ denotes the sigmoid function and
$\odot$ denotes element-wise multiplication.

For cross-sample augmentation, two copies of original network \(\mathbf{G}\) are permuted from \( \mathbf{R}^{B \times T \times N} \) to \( \mathbf{R}^{N \times T \times B} \) and \( \mathbf{R}^{T \times B \times N} \), respectively. The same Z-pool operation is applied over the last two dimensions of each permuted tensor, enabling batch--temporal and batch--spatial interactions. The permutation changes the tensor layout, while spatial graph convolution is applied along the ROI axis using the corresponding subject's adjacency matrix. Temporal convolution operates along the time axis. Each pooled representation is processed by the ST-GC encoder and multiplied element-wise by sigmoid output. The corresponding inverse permutation restores the branch outputs to the original \( \mathbf{R}^{B \times T \times N} \) ordering, and their equal-weight average forms \(\mathbf{G}^{\text{CA}}\). The resulting self-aware and cross-sample views are subsequently refined by the DAU module.

\begin{figure*}[!t]
    \centering
    \includegraphics[width=\textwidth]{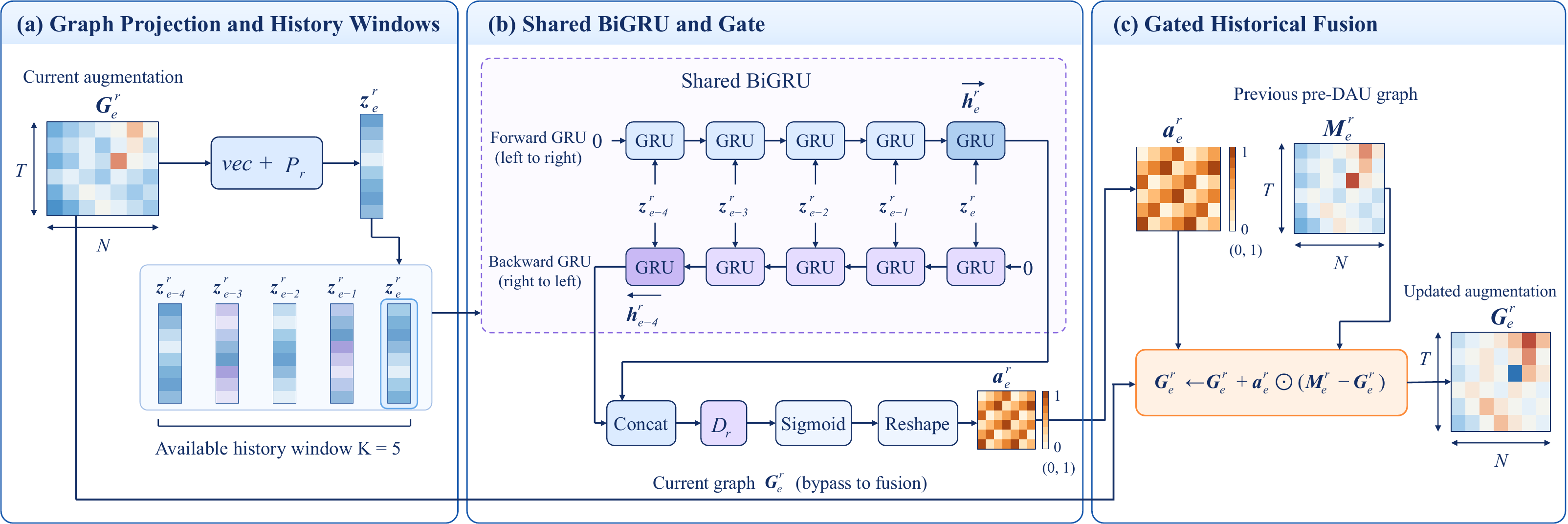}

    \caption{Illustration of the DAU module. (a) Whole-graph projection and construction of a descriptor history window across training epochs. (b) Shared BiGRU encoding followed by decoding and sigmoid gate generation. (c) Gated fusion of the current augmentation with the preceding epoch's pre-refinement graph. The self-aware and cross-sample branches share BiGRU parameters while maintaining separate histories, projections, and decoders.}
    \label{fig:DAU module}
    \vspace{-1.5em}
\end{figure*}

\subsubsection{Dynamic Augmentation Update Module}
\label{sec:DAU}

We introduce a dynamic augmentation update (DAU) module to refine augmented brain networks using historical information across training epochs, as shown in Fig.~\ref{fig:DAU module}. Let $\mathbf{G}_{e}^{r}$ denote the newly generated augmented network at epoch $e$, where $r\in\{\text{self-aware},\text{cross-sample}\}$ identifies the augmentation branch. A learnable projection $P_r:\mathbf{R}^\text{TN}\rightarrow\mathbf{R}^{d}$ maps the complete augmented matrix to a $d$-dimensional descriptor:
\begin{equation}
\mathbf{z}_{e}^{r}=P_r\left(\operatorname{vec}\left(\mathbf{G}_{e}^{r}\right)\right),
\label{eq:DAU_descriptor}
\end{equation}
The operator $\operatorname{vec}$ flattens the entire matrix into a vector. For each subject and augmentation branch, we organize the descriptors into a sliding window with a maximum length of $K\geq1$, containing the current epoch and up to $K-1$ preceding epochs:
\begin{equation}
\begin{aligned}
\mathcal{Z}_{e}^{r}&=\left[\mathbf{z}_{\ell_e}^{r},\ldots,\mathbf{z}_{e}^{r}\right], \quad
\ell_e=\max(1,e-K+1),
\end{aligned}
\label{eq:DAU_history_window}
\end{equation}

The index $\ell_e$ identifies the earliest epoch in the available window. A shared BiGRU processes the self-aware and cross-sample descriptor sequences separately, using the same parameters across branches while maintaining separate hidden states. Its forward and backward recurrences are defined as follows:
\begin{equation}
\begin{aligned}
\overrightarrow{\mathbf{h}_{j}^{r}}&=\overrightarrow{\operatorname{GRU}}\left(\mathbf{z}_{j}^{r},\overrightarrow{\mathbf{h}_{j-1}^{r}}\right), \quad
\overleftarrow{\mathbf{h}_{j}^{r}}=\overleftarrow{\operatorname{GRU}}\left(\mathbf{z}_{j}^{r},\overleftarrow{\mathbf{h}_{j+1}^{r}}\right),
\end{aligned}
\label{eq:DAU_bidirectional_recurrence}
\end{equation}
The index $j$ ranges from $\ell_e$ to $e$. The arrows indicate the traversal directions, and each hidden state is an $m$-dimensional vector. The histories and hidden states remain branch-specific despite parameter sharing. The forward boundary state at $\ell_e-1$ and the backward boundary state at $e+1$ are initialized to zero. Both directions process only the available window and require no representations from future epochs.

The forward state at the current endpoint and the backward state at the earliest endpoint summarize the window. A learnable decoder $D_r:\mathbf{R}^{2m}\rightarrow\mathbf{R}^{\text{TN}}$ maps their concatenation to an element-wise fusion gate. Let $\mathbf{M}_{e}^{r}\in\mathbf{R}^{\text{T}\times \text{N}}$ denote the historical reference, defined as the pre-refinement augmented graph from the preceding epoch for the same subject and branch. The reference and current graph share the same temporal and ROI coordinates, with $\mathbf{M}_{1}^{r}=\mathbf{G}_{1}^{r}$ in the first epoch. The gate and historical fusion are defined as follows:
\begin{equation}
\begin{gathered}
\mathbf{a}_{e}^{r}
=\operatorname{\text{reshape}}_{\text{T}\times \text{N}}\!\left\{\sigma\!\left(D_r\left[\overrightarrow{\mathbf{h}_{e}^{r}};\overleftarrow{\mathbf{h}_{\ell_e}^{r}}\right]\right)\right\},\\
\mathbf{G}_{e}^{r}
\leftarrow\mathbf{G}_{e}^{r}+\mathbf{a}_{e}^{r}\odot\left(\mathbf{M}_{e}^{r}-\mathbf{G}_{e}^{r}\right).
\end{gathered}
\label{eq:DAU_gated_fusion}
\end{equation}
Here, the semicolon denotes concatenation, $\sigma$ is the element-wise sigmoid function, and $\operatorname{\text{reshape}}_{\text{T}\times \text{N}}$ restores the matrix using the same ordering as $\operatorname{vec}$. The gate $\mathbf{a}_{e}^{r}\in(0,1)^{T\times N}$ controls the contribution of the historical reference, and $\odot$ denotes element-wise multiplication. The right-hand side uses the current graph before the update. Smaller gate values favor the current graph, whereas larger values favor the historical reference. Each updated entry lies between its current and historical values, without an additional global scaling hyperparameter. The first epoch produces an identity update.

\subsubsection{Inference for Test Subjects} 
At the end of training, we retain the model checkpoints from the last \(K\) consecutive epochs, including the final epoch \(E\). For each unseen subject \(i\), the same input brain network and connectivity matrix are processed by the augmentation modules from checkpoints \(j=E-K+1,\ldots,E\). This produces a subject-specific sequence of pre-DAU augmented graphs \(G_{i,j}^{r}\) for each branch \(r\in\{\mathrm{SA},\mathrm{CA}\}\). Each graph is mapped to a descriptor using the projection from its corresponding checkpoint, \(z_{i,j}^{r}=P_j^{r}(\operatorname{vec}(G_{i,j}^{r}))\). The descriptors are arranged in chronological checkpoint order to construct the history window \(Z_{i,E}^{r}\). Thus, an augmentation history is constructed separately for each unseen subject using the saved model checkpoints.

The shared BiGRU and branch-specific decoders from the final checkpoint \(\Theta_E\) process these histories to compute the fusion gates, with the recurrent boundary states initialized to zero for each subject and branch. Following Eq. (7), the current augmented graph \(G_{i,E}^{r}\) is refined using the preceding checkpoint’s pre-DAU graph \(G_{i,E-1}^{r}\) as the historical reference. The classification branch from \(\Theta_E\) then processes the original network and the two refined augmented views, and their predicted probabilities are averaged according to Eq. (14).

All checkpoints are frozen and used in evaluation mode, without test-time parameter updates or access to test labels. The same history-construction and prediction procedure is applied during validation and testing.

\subsection{Topology-guided Multi-view Graph Contrastive Learning}

We introduce a topology-guided multi-view graph contrastive
learning scheme to jointly constrain the relationships among
the original network and its two augmented views.
Specifically, the objective \( \mathcal{L}_{\text{TMC}} \) combines a consistency constraint
between the original network and the self-aware augmented view
with diversity constraints involving the cross-sample
augmented view.
This asymmetric design assigns consistency and diversity
roles to the two augmentation branches.

\subsubsection{Topology-guided Multi-view Graph Contrastive Loss}

We perform multi-view graph contrastive learning by leveraging the semantic information of the topological structures in augmented brain networks. First, we use cosine similarity~\cite{tang2022contrastive} to quantify
the relationships among the three views.
Taking the original brain network $\mathbf{G}$ and its
self-aware augmented view $\mathbf{G}^{\text{SA}}$ as an
example, the similarity is defined as:
\begin{equation}
\label{eq:cosine_similarity}
\operatorname{sim}(\mathbf{G},\mathbf{G}^\text{SA})
=
\frac{
    \left\langle \mathbf{G},\mathbf{G}^{\text{SA}}\right\rangle_{T,N}
}{
    \max\!\left(\|\mathbf{G}\|_{T,N},\varepsilon\right)
    \max\!\left(\|\mathbf{G}^{\text{SA}}\|_{T,N},\varepsilon\right)
},
\end{equation}
Here, $\langle\cdot,\cdot\rangle_{T,N}$ and
$\|\cdot\|_{T,N}$ denote the inner product and Euclidean norm
computed over the temporal and ROI dimensions, respectively,
with the batch dimension preserved. The normalization and division are performed element-wise over the preserved batch dimension. Specifically, the inner product is defined as:
\begin{equation}
\left\langle
    \mathbf{G},\mathbf{G}^{\text{SA}}
\right\rangle_{T,N}
=
\sum_{t=1}^{T}
\sum_{n=1}^{N}
\mathbf{G}_{:,t,n}
\odot
\mathbf{G}^{\text{SA}}_{:,t,n},
\end{equation}
The colon denotes the preserved batch dimension, and $\odot$ denotes element-wise multiplication. The constant $\varepsilon$ ensures numerical stability. A higher similarity score indicates greater similarity between the original network and its self-aware augmented view. 

The two augmentation branches are assigned different regularization roles.
For the self-aware view $\mathbf{G}^{\text{SA}}$, we use the original network $\mathbf{G}$ as a consistency reference. Encouraging agreement between these matched views is intended to limit augmentation-induced deviations in representation direction and maintain a subject-level correspondence with the original network.
Accordingly, $\mathcal{L}_{\text{SA}}$ favors stronger matching relative to other pairings within the mini-batch.

The cross-sample branch applies distinct transformations to spatial and temporal axes to generate an alternative view of the original network.
Requiring this view to closely match the original representation may suppress the differences introduced by these transformations.
We therefore use $\mathcal{L}_{\text{CA}}$ to discourage the matched association relative to other pairings within the mini-batch, with the aim of retaining view diversity.
The two constraints are formulated as follows:
\begin{equation}
\begin{aligned}
\mathcal{L}_{\text{SA}}
=-\frac{1}{B}\sum_{i=1}^{B}
\log\frac{\exp\!\left(\operatorname{sim}(\mathbf{G}_{i}^{\text{SA}},\mathbf{G}_i)/\tau\right)}
{\sum_{j=1}^{B}\exp\!\left(\operatorname{sim}(\mathbf{G}_{i}^{\text{SA}},\mathbf{G}_j)/\tau\right)},
\end{aligned}
\label{eq:L_SA}
\end{equation}
\begin{equation}
\begin{aligned}
\mathcal{L}_{\text{CA}}=\frac{1}{B}\sum_{i=1}^{B}\log\frac{\exp\!\left(\operatorname{sim}(\mathbf{G}_{i}^{\text{CA}},\mathbf{G}_i)/\tau\right)}
{\sum_{j=1}^{B}\exp\!\left(\operatorname{sim}(\mathbf{G}_{i}^{\text{CA}},\mathbf{G}_j)/\tau\right)},
\end{aligned}
\label{eq:L_CA}
\end{equation}
where $\tau>0$ is the temperature parameter, set to 0.5 in our experiments.
The constraints involving the original network do not directly regulate the similarity between the two augmented views~\cite{xu2024graph}.
In particular, reducing the cross-sample to original matching score alone does not ensure that the cross-sample branch expresses a perspective distinct from the self-aware branch.
We therefore introduce $\mathcal{L}_{\text{SA,CA}}$ to explicitly discourage strong matched associations between the augmented views relative to other pairings within the mini-batch.
This additional constraint is intended to reduce overlap between the two augmentation branches and complement their respective relationships with the original network.
The inter-view constraint is formulated as follows:
\begin{equation}
\begin{aligned}
\mathcal{L}_{\text{SA,CA}}=\frac{1}{B}\sum_{i=1}^{B}\log\frac{\exp\!\left(\operatorname{sim}(\mathbf{G}_{i}^{\text{SA}},\mathbf{G}_{i}^{\text{CA}})/\tau\right)}
{\sum_{j=1}^{B}\exp\!\left(\operatorname{sim}(\mathbf{G}_{i}^{\text{SA}},\mathbf{G}_{j}^{\text{CA}})/\tau\right)},
\end{aligned}
\label{eq:L_SA_CA}
\end{equation}

Combining the three pairwise constraints yields the multi-view graph contrastive loss: $\mathcal{L}_{\text{TMC}}$:
\begin{equation}
\begin{aligned}
\mathcal{L}_{\text{TMC}}
&=\alpha\mathcal{L}_{\text{SA}}+(1-\alpha)\mathcal{L}_{\text{CA}}+\beta\mathcal{L}_{\text{SA,CA}}.
\end{aligned}
\label{eq:TMC loss}
\end{equation}
where $\alpha \in [0,1]$ balances the consistency and diversity constraints, and $\beta \geq 0$ controls the contribution of the constraint between the two augmented views.

Together, these constraints combine an original-view consistency reference with two diversity-oriented relationships.
They are used as similarity-based regularizers alongside the supervised classification objective.
The optimization directions of these constraints can be characterized through their partial derivatives with respect to the similarity scores.
For $B \geq 2$ and $\tau>0$, the negative log-ratio in $\mathcal{L}_{\text{SA}}$ has a negative derivative with respect to the matched similarity score, whereas the positive log-ratios in $\mathcal{L}_{\text{CA}}$ and $\mathcal{L}_{\text{SA,CA}}$ have positive derivatives.
Thus, at the similarity-score level, minimizing these terms favors increasing the matched self-aware with original similarity and decreasing the matched cross-sample with original and self-aware with cross-sample similarities.
The detailed derivation and the corresponding effects on non-matched pairs are provided in~\ref{app:optimization_direction}.

\subsubsection{Objective Function}

For downstream brain disorder diagnosis, we employ a composite objective that combines the classification loss $\mathcal{L}_{\text{class}}$ and the contrastive loss $\mathcal{L}_{\text{TMC}}$. Specifically, the original brain networks $\mathbf{G}$ and the two augmented brain networks $\mathbf{G}^{\text{SA}}$ and $\mathbf{G}^{\text{CA}}$ are processed by the same classification branch. Each network is first passed through the ST-GC encoder to extract spatio-temporal features. These features are averaged over the ROI dimension and flattened into feature vectors, which are then fed into a shared classification head consisting of four successive $1\times1$ convolutional layers. The scalar output for each view is transformed by a sigmoid function to obtain its predicted patient probability. The three probabilities are averaged to obtain the final prediction, which is supervised using the ground-truth diagnostic label:
\begin{equation}
\begin{gathered}
p_i=\frac{1}{3}\left(p_i^{\text{orig}}+p_i^{\text{SA}}+p_i^{\text{CA}}\right),\\
\mathcal{L}_{\text{class}}=-\sum_{i=1}^{B}\left[y_i\log(p_i)+(1-y_i)\log(1-p_i)\right].
\end{gathered}
\label{eq:classification loss}
\end{equation}
Here, $p_i^{\text{orig}}$, $p_i^{\text{SA}}$, and $p_i^{\text{CA}}$ denote the predicted patient probabilities for the original, self-aware, and cross-sample views of subject $i$, respectively. $B$ is the mini-batch size, and $y_i\in\{0,1\}$ is the ground-truth label, with $1$ indicating a patient and $0$ indicating a healthy control. The classification loss provides diagnostic supervision through the averaged prediction, while $\mathcal{L}_{\text{TMC}}$ regularizes the relationships among the three views through the consistency and diversity constraints described above.
The overall training objective is:
\begin{equation}
\mathcal{L} = \mathcal{L}_{\text{class}} + \lambda \ \mathcal{L}_{\text{TMC}}.
\label{eq:loss}
\end{equation}
where $\lambda \geq 0$ controls the contribution of the
multi-view contrastive regularization relative to the
classification loss.

\begin{algorithm}[!t]
\small
\caption{Training Procedure of MCDA}
\label{alg:MCDA}
\textbf{Input}: Brain networks $\mathbf{G}$, connectivity matrices $\mathbf{A}$, and labels $y$ \\
\textbf{Output}: Trained model parameters $\Theta$ \\
\textbf{Parameters}: Epochs $E$, batch size $B$, history length $K$, temperature $\tau$, and loss weights $\lambda$, $\alpha$, $\beta$

\begin{algorithmic}[1]
\STATE Initialize $\Theta$, Adam, and subject- and branch-specific histories
\FOR{$e=1$ to $E$}
    \FOR{each training mini-batch}
        \STATE $\mathbf{G}_{e}^{\text{SA}}\gets\operatorname{Z-pool}(\mathbf{G})\odot\sigma\!\left(\operatorname{ST-GC}\left(\operatorname{Z-pool}(\mathbf{G}),\mathbf{A}\right)\right)$
        \STATE $\mathbf{G}_{2,z}\gets\operatorname{Z-pool}(\operatorname{Rotate}_{N}(\mathbf{G}))$
        \STATE $\mathbf{G}_{2,s}\gets\operatorname{Rotate}_{N}^{-1}[\mathbf{G}_{2,z}\odot\sigma(\operatorname{ST-GC}(\mathbf{G}_{2,z},\mathbf{A}))]$
        \STATE $\mathbf{G}_{3,z}\gets\operatorname{Z-pool}(\operatorname{Rotate}_{T}(\mathbf{G}))$
        \STATE $\mathbf{G}_{3,t}\gets\operatorname{Rotate}_{T}^{-1}[\mathbf{G}_{3,z}\odot\sigma(\operatorname{ST-GC}(\mathbf{G}_{3,z},\mathbf{A}))]$
        \STATE $\mathbf{G}_{e}^{\text{CA}}\gets\frac{1}{2}\left(\mathbf{G}_{2,s}+\mathbf{G}_{3,t}\right)$
        \FOR{each subject and $r\in\{\text{SA},\text{CA}\}$}
            \STATE $\begin{aligned}[t]
                \mathbf{z}_{e}^{r}&\gets P_r(\operatorname{vec}(\mathbf{G}_{e}^{r})),\\
                \mathcal{Z}_{e}^{r}&\gets[\mathbf{z}_{\ell_e}^{r},\ldots,\mathbf{z}_{e}^{r}]
            \end{aligned}$
            \STATE Retain detached copies of $(\mathbf{z}_{e}^{r},\mathbf{G}_{e}^{r})$ for caching
            \STATE $\left(\overrightarrow{\mathbf{h}_{e}^{r}},\overleftarrow{\mathbf{h}_{\ell_e}^{r}}\right)\gets\operatorname{BiGRU}(\mathcal{Z}_{e}^{r};\mathbf{0})$
            \STATE $\mathbf{a}_{e}^{r}\gets\operatorname{\text{reshape}}_{T\times N}\left(\sigma\left(D_r\left[\overrightarrow{\mathbf{h}_{e}^{r}};\overleftarrow{\mathbf{h}_{\ell_e}^{r}}\right]\right)\right)$
            \STATE $\mathbf{G}_{e}^{r}\gets\mathbf{G}_{e}^{r}+\mathbf{a}_{e}^{r}\odot(\mathbf{M}_{e}^{r}-\mathbf{G}_{e}^{r}),\quad \text{if }e>1$
        \ENDFOR
        \STATE $p \gets \frac{1}{3}\left(p^{\text{orig}} + p^{\text{SA}} + p^{\text{CA}}\right),$
        \STATE $\mathcal{L}_{\text{TMC}}\gets\alpha\mathcal{L}_{\text{SA}}+(1-\alpha)\mathcal{L}_{\text{CA}}+\beta\mathcal{L}_{\text{SA,CA}}$
        \STATE $\mathcal{L}\gets\mathcal{L}_{\text{class}}+\lambda\mathcal{L}_{\text{TMC}}$
        \STATE $\Theta\gets\operatorname{\text{Adam}}(\Theta,\nabla_{\Theta}\mathcal{L})$
        \STATE Update the corresponding histories using the retained copies
    \ENDFOR
\ENDFOR
\RETURN $\Theta$
\end{algorithmic}
\end{algorithm}

\subsection{Implementation Details}
\label{sec:implementation details}

The framework is implemented using PyTorch~\cite{paszke2017automatic} on an NVIDIA RTX 4090 GPU. The local neighborhood size in Z-pool is set to $|R|$=16, corresponding to a 4$\times$4 region. The ST-GC encoders contain three layers. The maximum DAU history window length is set to $K$=5, including the current epoch and up to four preceding epochs. The influence of the history window length is examined in Section~\ref{sec:influece of historical windows size}. Each branch-specific projection $P_r$ is implemented as a single fully connected layer mapping the flattened augmented graph to a descriptor of dimension $d$=128. The shared BiGRU contains three layers, with a hidden-state dimension of $m$=64 per direction and an inter-layer dropout rate of 0.3. The terminal states from the highest layer are concatenated into a 128-dimensional vector. Each branch-specific decoder $D_r$ is a single fully connected layer mapping this vector to $\text{TN}$ gate logits, followed by the sigmoid and reshape operations defined in Eq.~\eqref{eq:DAU_gated_fusion}. Both fully connected layers include biases.

All trainable modules are initialized before constructing the Adam optimizer~\cite{kingma2014adam}. Training runs for 200 epochs with an initial learning rate of 0.001, which is multiplied by 0.5 every 50 epochs. The learning-rate scheduler is updated at the end of each epoch. The batch size is specified in Section~\ref{sec:influece of batch size}, and the loss weights are set to $\alpha=0.3$, $\beta=1.0$, and $\lambda=0.5$.
During training, the original brain network and its self-aware and cross-sample augmented views are processed by the same classification branch. Their predicted patient probabilities are averaged according to Eq. (14), and the averaged probability is used to compute the classification loss.
At test time, we construct augmentation histories for each unseen subject using the last \(K\) epoch-end checkpoints. Each checkpoint generates pre-DAU views and descriptors using its augmentation modules and projections. The chronologically ordered descriptors are processed by the final checkpoint’s DAU, using the preceding checkpoint’s pre-DAU views as references in Eq. (7). The final checkpoint’s classifier processes the original and two refined views, and their predicted probabilities are averaged according to Eq. (14). All checkpoints are frozen and used in evaluation mode. For clarity, we show the detailed training process of our proposed MCDA framework in Algorithm~\ref{alg:MCDA}.

% Each training subject is processed once per epoch using shuffled mini-batches. Historical information is indexed by subject identity and augmentation branch rather than by mini-batch position. Each history retains up to $K-1$ preceding descriptors and the most recent pre-refinement augmented graph. Historical descriptors and reference graphs are detached from the computational graph, while the current descriptor remains differentiable. The BiGRU hidden states are initialized to zero for each window rather than carried between mini-batches. Before historical fusion, copies of the current descriptor and pre-refinement graph are captured from the forward pass. After the optimizer update, these copies are detached and stored for subsequent epochs without recomputation. Descriptor entries exceeding the history length are discarded, and the stored reference graph is replaced by the current pre-refinement snapshot. Historical fusion is bypassed in the first epoch, when no preceding reference is available.

\section{Experiment}\label{sec:experiment}

\subsection{Materials and Data Pre-processing}

\begin{table}[!t]
\centering
\footnotesize
\setlength{\tabcolsep}{2pt}
\renewcommand{\arraystretch}{1.06}
\caption{Demographic Information of studied subjects from the REST-MDD and ABIDE datasets. The age is reported as mean $\pm$ standard deviation. M: Male; F: Female; MDD: Major depressive disorder; ASD: Autism Spectrum Disorder; HC: Healthy control.}
\label{tab:dataset_info}
\begin{tabular*}{\columnwidth}{@{\extracolsep{\fill}}lccccc@{}}
\toprule
\textbf{Dataset} & \textbf{Site} & \textbf{Category} & \textbf{Number} & \shortstack{\textbf{Gender}\textbf{(M/F)}} & \textbf{Age} \\
\midrule
\multirow{6}{*}{REST-MDD} 
    & \multirow{2}{*}{20} & MDD & 282 & 99/183 & 38.74 $\pm$ 13.74 \\
    &                     & HC  & 251 & 87/164 & 39.64 $\pm$ 15.87 \\ 
    & \multirow{2}{*}{21} & MDD & 86  & 38/48  & 34.71 $\pm$ 12.63 \\
    &                     & HC  & 70  & 31/39  & 36.13 $\pm$ 12.64 \\
    & \multirow{2}{*}{25} & MDD & 89  & 21/68  & 65.60 $\pm$ 6.75 \\
    &                     & HC  & 63  & 29/34  & 69.63 $\pm$ 5.86 \\
\midrule
\multirow{6}{*}{ABIDE} 
    & \multirow{2}{*}{NYU}    & ASD & 79  & 68/11 & 14.52 $\pm$ 6.97 \\
    &                         & HC  & 105 & 79/26 & 15.81 $\pm$ 6.25 \\
    & \multirow{2}{*}{UM}     & ASD & 55  & 46/9  & 12.72 $\pm$ 2.40 \\
    &                         & HC  & 55  & 38/17 & 14.07 $\pm$ 3.18 \\
    & \multirow{2}{*}{USM}    & ASD & 58  & 58/0  & 22.65 $\pm$ 7.73 \\
    &                         & HC  & 43  & 43/0  & 21.36 $\pm$ 7.64 \\
\bottomrule
\vspace{-2.5em}
\end{tabular*}
\end{table}

We evaluate the proposed MCDA framework on two rs-fMRI datasets: 1) REST-meta-MDD Consortium (REST-MDD\footnote{\url{http://rfmri.org/REST-meta-MDD}})~\cite{yan2019reduced} and 2) Autism Brain Imaging Data Exchange (ABIDE\footnote{\url{http://fcon_1000.projects.nitrc.org/indi/abide}})~\cite{di2014autism}. 
For each dataset, we use the top three largest sites in the experiments. The demographic information of these subjects from REST-MDD and ABIDE is provided in Table~\ref{tab:dataset_info}. 

\textbf{1) REST-MDD.} For the REST-MDD dataset, the three largest sites are selected for analysis: Site 20, Site 21, and Site 25, which respectively include 282 patients with major depressive disorder (MDD) and 251 healthy controls (HCs); 86 MDD patients and 70 HCs; and 89 MDD patients and 63 HCs, with acquisition details provided in the original dataset publication~\cite{yan2019reduced}. All rs-fMRI data from REST-MDD were preprocessed using the DPARSF pipeline~\cite{yan2010dparsf}. The first 10 volumes were discarded, followed by standard preprocessing procedures including head motion correction, spatial smoothing, normalization, band-pass filtering ($\text{0.01--0.10,Hz}$), nuisance signal regression, and spatial standardization.

\textbf{2) ABIDE.} In ABIDE, we use three sites, including Site NYU with 79 ASD and 105 HC subjects, Site UM with 55 ASD and 55 HC subjects, and Site USM with 58 ASD and 43 HC subjects, with acquisition details provided in the original dataset publication~\cite{di2014autism}. The rs-fMRI data were preprocessed based on the DPARSF pipeline~\cite{yan2010dparsf}.
% , which involved head motion correction, nuisance covariates regression (ventricle, CSF, WM, and higher-order effects of head motion), spatial normalization to the MNI template, and temporal high-pass filtering (0.01--0.10 Hz) using a linear detrending and fast Fourier transform.

For both datasets, the brain was parcellated into 116 ROIs using the original Automated Anatomical Labeling (AAL) atlas~\cite{tzourio2002automated}, and the mean BOLD time series was extracted from each ROI. The repetition time (TR) was 2000~ms at all six sites. For each subject, the final model input consisted of 200 time points from 116 ROIs, yielding $\mathbf{G}\in\mathbf{R}^{200\times116}$.

\subsection{Experiment Setting}

\begin{table*}[!t]
\centering
\footnotesize
\caption{Results (\%) of seven methods in MDD vs. HC classification with mean $\pm$ standard deviation. Best results are shown in bold. Here, * denotes a statistically significant (\textit{p}\textless 0.05) difference between MCDA and a competing method in terms of accuracy.}
\label{tab:classifciation experiments of MDD}

\setlength{\tabcolsep}{1.25pt}

\begin{tabular*}{\textwidth}{@{\extracolsep{\fill}}lcccccccccccc@{}}
\toprule
\multirow{2}{*}{Method}
& \multicolumn{4}{c}{Site 20}
& \multicolumn{4}{c}{Site 21}
& \multicolumn{4}{c}{Site 25} \\
\cmidrule(lr){2-5} \cmidrule(lr){6-9} \cmidrule(lr){10-13}
& ACC & AUC & PRE & REC
& ACC & AUC & PRE & REC
& ACC & AUC & PRE & REC \\
\midrule
BrainGNN &
\tabresult{63.6}{1.7}{*} & \tabresult{65.5}{3.3}{} & \tabresult{57.6}{5.9}{} & \textbf{\tabresult{69.8}{14.2}{}} &
\tabresult{65.2}{6.2}{} & \tabresult{67.0}{2.9}{} & \tabresult{58.8}{7.2}{} & \textbf{\tabresult{74.2}{7.5}{}} &
\tabresult{66.2}{8.3}{} & \tabresult{64.8}{7.6}{} & \tabresult{59.3}{7.2}{} & \tabresult{68.0}{8.2}{} \\
LCAF &
\tabresult{61.6}{1.9}{*} & \tabresult{62.8}{4.0}{} & \tabresult{54.3}{2.7}{} & \tabresult{68.9}{7.3}{} &
\tabresult{61.1}{2.9}{*} & \tabresult{55.5}{6.6}{} & \tabresult{55.3}{10.1}{} & \tabresult{57.3}{12.7}{} &
\tabresult{69.0}{7.3}{} & \tabresult{67.6}{3.6}{} & \tabresult{63.0}{6.8}{} & \tabresult{70.0}{6.1}{} \\
TSDA &
\tabresult{62.1}{2.4}{*} & \tabresult{60.9}{4.6}{} & \tabresult{56.1}{3.9}{} & \tabresult{62.9}{7.8}{} &
\tabresult{59.0}{2.7}{*} & \tabresult{55.0}{5.8}{} & \tabresult{52.6}{7.9}{} & \tabresult{62.9}{17.0}{} &
\tabresult{62.7}{5.7}{*} & \tabresult{54.2}{4.1}{} & \tabresult{52.1}{1.1}{} & \tabresult{63.8}{13.9}{} \\
GCDA &
\tabresult{63.5}{1.0}{*} & \tabresult{62.2}{4.0}{} & \tabresult{59.4}{4.2}{} & \tabresult{64.5}{14.3}{} &
\tabresult{67.3}{4.1}{} & \tabresult{65.2}{6.5}{} & \tabresult{60.0}{10.5}{} & \tabresult{69.4}{7.8}{} &
\tabresult{69.3}{3.3}{} & \tabresult{68.8}{6.3}{} & \textbf{\tabresult{70.7}{7.6}{}} & \tabresult{72.6}{9.6}{} \\
UCGL &
\tabresult{63.7}{2.7}{*} & \tabresult{60.7}{2.7}{} & \tabresult{57.9}{2.9}{} & \tabresult{66.2}{11.1}{} &
\tabresult{62.8}{3.3}{*} & \tabresult{61.2}{5.7}{} & \tabresult{52.0}{8.3}{} & \tabresult{63.4}{13.1}{} &
\tabresult{63.3}{1.9}{*} & \tabresult{58.0}{4.4}{} & \tabresult{57.1}{9.8}{} & \tabresult{67.2}{12.3}{} \\
MCS-GL &
\tabresult{64.3}{3.7}{*} & \tabresult{65.3}{4.0}{} & \tabresult{61.1}{6.1}{} & \tabresult{63.9}{14.1}{} &
\tabresult{69.0}{5.4}{} & \tabresult{68.5}{6.3}{} & \tabresult{62.5}{14.2}{} & \tabresult{69.8}{9.8}{} &
\tabresult{70.6}{3.4}{*} & \tabresult{69.5}{5.6}{} & \tabresult{69.2}{5.7}{} & \tabresult{72.9}{11.4}{} \\
\midrule
MCDA &
\textbf{\tabresult{70.3}{2.9}{}} & \textbf{\tabresult{74.5}{2.8}{}} & \textbf{\tabresult{72.3}{4.0}{}} & \tabresult{65.6}{8.4}{} &
\textbf{\tabresult{71.8}{6.4}{}} & \textbf{\tabresult{70.2}{5.3}{}} & \textbf{\tabresult{63.9}{4.8}{}} & \tabresult{69.5}{8.8}{} &
\textbf{\tabresult{73.4}{3.4}{}} & \textbf{\tabresult{73.8}{2.5}{}} & \tabresult{66.6}{6.5}{} & \textbf{\tabresult{79.4}{9.8}{}} \\
\bottomrule
\end{tabular*}
\end{table*}

\begin{table*}[!t]
\centering
\footnotesize
\caption{Results (\%) of seven methods in ASD vs. HC classification with mean $\pm$ standard deviation. Best results are shown in bold. Here, * denotes a statistically significant (\textit{p}\textless 0.05) difference between MCDA and a competing method in terms of accuracy.}
\label{tab:classification experiments of ASD}

\setlength{\tabcolsep}{1.25pt}

\begin{tabular*}{\textwidth}{@{\extracolsep{\fill}}lcccccccccccc@{}}
\toprule
\multirow{2}{*}{Method}
& \multicolumn{4}{c}{Site NYU}
& \multicolumn{4}{c}{Site UM}
& \multicolumn{4}{c}{Site USM} \\
\cmidrule(lr){2-5} \cmidrule(lr){6-9} \cmidrule(lr){10-13}
& ACC & AUC & PRE & REC
& ACC & AUC & PRE & REC
& ACC & AUC & PRE & REC \\
\midrule
BrainGNN &
\tabresult{73.6}{7.8}{*} & \tabresult{70.3}{5.9}{} & \tabresult{70.3}{5.8}{} & \tabresult{49.9}{7.5}{} &
\tabresult{70.9}{6.1}{} & \tabresult{70.4}{5.6}{} & \tabresult{81.0}{14.9}{} & \tabresult{51.3}{14.2}{} &
\tabresult{75.2}{9.1}{*} & \tabresult{71.1}{10.2}{} & \tabresult{69.2}{13.9}{} & \tabresult{84.7}{11.5}{} \\
LCAF &
\tabresult{71.7}{4.6}{*} & \tabresult{72.2}{4.8}{} & \tabresult{77.6}{9.2}{} & \tabresult{64.9}{12.4}{} &
\tabresult{71.2}{5.2}{} & \tabresult{73.0}{5.7}{} & \tabresult{72.2}{13.4}{} & \tabresult{61.8}{8.9}{} &
\tabresult{71.5}{5.1}{*} & \tabresult{73.9}{11.2}{} & \tabresult{70.6}{15.9}{} & \tabresult{73.5}{12.4}{} \\
TSDA &
\tabresult{68.3}{7.0}{*} & \tabresult{69.3}{7.3}{} & \tabresult{73.1}{9.8}{} & \tabresult{66.6}{8.5}{} &
\tabresult{69.5}{4.2}{*} & \tabresult{70.8}{9.9}{} & \tabresult{72.5}{17.7}{} & \tabresult{68.8}{15.9}{} &
\tabresult{62.4}{4.5}{*} & \tabresult{57.4}{5.4}{} & \tabresult{66.1}{6.4}{} & \tabresult{69.2}{2.2}{} \\
GCDA &
\tabresult{76.4}{2.4}{} & \tabresult{65.7}{6.0}{} & \tabresult{76.2}{8.3}{} & \tabresult{66.7}{15.6}{} &
\tabresult{71.2}{13.4}{} & \tabresult{74.5}{14.8}{} & \tabresult{65.3}{12.1}{} & \textbf{\tabresult{77.9}{16.2}{}} &
\tabresult{77.8}{8.6}{} & \tabresult{73.9}{14.1}{} & \textbf{\tabresult{81.7}{6.9}{}} & \tabresult{78.6}{15.1}{} \\
UCGL &
\tabresult{73.0}{6.6}{*} & \tabresult{70.7}{8.2}{} & \textbf{\tabresult{78.9}{8.2}{}} & \tabresult{59.6}{17.5}{} &
\tabresult{74.5}{5.2}{} & \tabresult{71.8}{7.7}{} & \tabresult{81.1}{6.5}{} & \tabresult{58.2}{18.7}{} &
\tabresult{77.2}{7.5}{} & \tabresult{78.6}{9.0}{} & \tabresult{78.0}{9.4}{} & \tabresult{78.1}{9.4}{} \\
MCS-GL &
\tabresult{70.8}{8.3}{*} & \tabresult{73.1}{6.2}{} & \tabresult{68.1}{6.4}{} & \tabresult{62.9}{12.2}{} &
\tabresult{73.9}{3.2}{} & \tabresult{76.7}{10.6}{} & \tabresult{76.5}{8.9}{} & \tabresult{77.3}{14.7}{} &
\tabresult{70.6}{6.5}{*} & \tabresult{72.6}{6.6}{} & \tabresult{71.1}{8.6}{} & \tabresult{79.9}{7.8}{} \\
\midrule
MCDA &
\textbf{\tabresult{81.1}{2.1}{}} & \textbf{\tabresult{82.9}{2.4}{}} & \tabresult{76.1}{9.9}{} & \textbf{\tabresult{68.1}{8.0}{}} &
\textbf{\tabresult{75.0}{2.6}{}} & \textbf{\tabresult{78.4}{7.2}{}} & \textbf{\tabresult{87.4}{3.6}{}} & \tabresult{61.2}{9.6}{} &
\textbf{\tabresult{80.2}{7.1}{}} & \textbf{\tabresult{82.1}{5.8}{}} & \tabresult{77.9}{8.9}{} & \textbf{\tabresult{86.2}{8.4}{}} \\
\bottomrule
\end{tabular*}
\vspace{-1em}
\end{table*}

In our experiments, we perform five-fold cross-validation independently at each site using random subject-level splits. In each fold, one subset is held out for testing, while the remaining data are split into training and validation sets at an 80:20 ratio for model fitting and selection, respectively. All competing methods are retrained using identical data partitions. Model selection is based solely on validation performance. Results are reported as the mean and standard deviation across the five test folds.
The evaluation metrics include accuracy (ACC), area under the receiver operating characteristic curve (AUC), precision (PRE), and recall (REC). 

We compare MCDA with the following six SOTA methods for fMRI analysis. Specifically, two methods (i.e., BrainGNN\footnote{\url{https://github.com/xxlya/BrainGNN_Pytorch}} and LCAF\footnote{\url{https://github.com/ku-milab/learnable-counter-condition-FC}}) are advanced rs-fMRI analysis approaches, followed by augmentation-based approaches (i.e., TSDA\footnote{\url{https://github.com/mxliu/ACTION-Software}} and GCDA) that enrich brain network representations. The remaining two methods (i.e., UCGL\footnote{\url{https://github.com/mxliu/Unsupervised-Contrastive-Graph-Learning}} and MCS-GL) adopt graph contrastive learning from different views for classification. 

\textbf{1) BrainGNN~\cite{li2021braingnn}:}
This is a graph neural network designed for analyzing fMRI and detecting neurological biomarkers. With the brain network/graph as input, BrainGNN uses two node-level graph convolutional layers to learn the node representations, capturing topological and functional patterns from fMRI data.

\textbf{2) LCAF~\cite{kang2023learnable}:}
Kang \textit{et al.} propose a unified FC-based framework integrating feature selection, feature extraction, and interpretation to reduce error propagation across stages. An adaptive attention network selects subject-specific, disease-related connections, while a functional network relational encoder models inter-network relationships to capture global topology without predefined edges. 

\textbf{3) TSDA~\cite{oh2020time}:}
In our implementation, TSDA generates two augmented sequences by extracting the first and last 90\% of each ROI's mean BOLD time series. Subsequent processing follows the same pipeline as MCDA for brain disorder classification.

\textbf{4) GCDA~\cite{wang2025self}:}
GCDA combines graph diffusion augmentation with two shared-parameter graph isomorphism networks for contrastive pretraining, followed by supervised fine-tuning. We adapt this two-stage procedure to our five-fold protocol: pretraining uses the training subjects without labels, while fine-tuning uses the same subjects with their diagnostic labels.

\textbf{5) UCGL~\cite{wang2023unsupervised}:}
UCGL employs dual-stage fMRI augmentation and two parallel ST-GCNs for contrastive pretraining, followed by supervised fine-tuning. We replace the fixed 60\%/20\%/20\% allocation with our fold-wise partitions, using the training subjects for both stages and their labels only during fine-tuning. For GCDA and UCGL methods, validation subjects are used only for model selection, and test subjects are reserved for final evaluation.

\textbf{6) MCS-GL~\cite{tian2024multi}:}
The MCS-GL framework enhances functional brain network-based diagnosis by expanding samples via Personalized PageRank (PPR) and incorporating local-global information as multi-view inputs for contrastive learning. During training, it extracts informative fMRI representations from unlabeled data to enable effective feature learning without supervision. In the final stage, model parameters are updated using a supervised cross-entropy loss.

\subsection{Classification Results}

The quantitative results of MCDA and the six competing methods for MDD vs. HC classification on the REST-MDD dataset and ASD vs. HC classification on the ABIDE dataset are reported in Table~\ref{tab:classifciation experiments of MDD} and Table~\ref{tab:classification experiments of ASD}, respectively, under within-site five-fold cross-validation. We perform two-sided paired $t$-tests on the ACC values of MCDA and each competing method obtained from the same five outer test folds ($n=5$ pairs; degrees of freedom $=4$). ``*'' denotes a nominal $p<0.05$ for the difference in ACC between MCDA and the corresponding method. From these tables, we have the following observations.

\emph{First}, compared to BrainGNN and LCAF, MCDA shows improved discriminative power. For instance, MCDA achieves ACC improvements of 6.7 and 6.6 percentage points over BrainGNN on Site 20 and Site 21 of REST-MDD, respectively. The possible reason is that topology-guided multi-view contrastive learning effectively aligns multiple views of brain connectivity by preserving topology semantics, enabling invariant and discriminative representation learning across views.

\emph{Second}, compared with the two augmentation-based methods, TSDA and GCDA, MCDA achieves higher mean ACC and AUC at each evaluated site. On Site NYU of ABIDE, MCDA exceeds TSDA by 12.8 percentage points in ACC and 13.6 percentage points in AUC. The performance gap suggests that simple window-slicing augmentation may be insufficient to capture the complex functional patterns needed for ASD diagnosis. In contrast, MCDA's dynamic topology augmentation maintains the brain's complete structures, allowing its contrastive learning to capture both localized dysfunctions and global network characteristics.

\emph{Additionally}, MCDA achieves higher mean ACC and AUC than the graph contrastive learning methods UCGL and MCS-GL at each evaluated site. Specifically, for MDD diagnosis on Site 25, MCDA exceeds UCGL by 10.1 percentage points in ACC and 15.8 percentage points in AUC. These results support the effectiveness of dynamic augmentation and topology-guided multi-view contrastive learning for within-site brain disorder classification.

\subsection{Ablation Study}

\begin{table*}[!t]
\centering
\footnotesize
\caption{Results (\%) of MCDA and its three variants in MDD vs. HC classification. Best results are shown in bold. Here, * denotes a statistically significant (\textit{p}\textless 0.05) difference between MCDA and a variant in terms of accuracy.}
\label{tab:ablation study of MDD}

\setlength{\tabcolsep}{1.25pt}

\begin{tabular*}{\textwidth}{@{\extracolsep{\fill}}lcccccccccccc@{}}
\toprule
\multirow{2}{*}{Method}
& \multicolumn{4}{c}{Site 20}
& \multicolumn{4}{c}{Site 21}
& \multicolumn{4}{c}{Site 25} \\
\cmidrule(lr){2-5} \cmidrule(lr){6-9} \cmidrule(lr){10-13}
& ACC & AUC & PRE & REC
& ACC & AUC & PRE & REC
& ACC & AUC & PRE & REC \\
\midrule
MCDA-I &
\tabresult{61.5}{2.6}{*} & \tabresult{50.9}{3.2}{} & \tabresult{53.8}{2.7}{} & \tabresult{63.8}{8.4}{} &
\tabresult{63.2}{5.4}{*} & \tabresult{65.4}{2.5}{} & \tabresult{54.0}{8.5}{} & \tabresult{66.6}{12.3}{} &
\tabresult{63.3}{1.9}{*} & \tabresult{58.8}{4.6}{} & \tabresult{52.7}{5.8}{} & \tabresult{63.8}{11.8}{} \\
MCDA-II &
\tabresult{60.8}{3.6}{*} & \tabresult{63.5}{7.3}{} & \tabresult{58.9}{4.5}{} & \tabresult{53.8}{10.7}{} &
\tabresult{68.0}{4.9}{*} & \tabresult{69.3}{6.5}{} & \tabresult{52.4}{11.6}{} & \tabresult{60.9}{19.0}{} &
\tabresult{68.0}{8.1}{} & \tabresult{64.1}{10.7}{} & \tabresult{59.5}{18.5}{} & \tabresult{54.1}{7.8}{} \\
MCDA-III &
\tabresult{58.2}{5.6}{*} & \tabresult{55.6}{6.5}{} & \tabresult{54.3}{4.9}{} & \tabresult{49.3}{8.9}{} &
\tabresult{65.4}{4.9}{*} & \tabresult{53.7}{12.5}{} & \tabresult{57.5}{10.4}{} & \tabresult{58.7}{13.1}{} &
\tabresult{64.4}{4.6}{*} & \tabresult{67.5}{5.4}{} & \textbf{\tabresult{68.5}{2.6}{}} & \tabresult{66.3}{3.2}{} \\
\midrule
MCDA &
\textbf{\tabresult{70.3}{2.9}{}} & \textbf{\tabresult{74.5}{2.8}{}} & \textbf{\tabresult{72.3}{4.0}{}} & \textbf{\tabresult{65.6}{8.4}{}} &
\textbf{\tabresult{71.8}{6.4}{}} & \textbf{\tabresult{70.2}{5.3}{}} & \textbf{\tabresult{63.9}{4.8}{}} & \textbf{\tabresult{69.5}{8.8}{}} &
\textbf{\tabresult{73.4}{3.4}{}} & \textbf{\tabresult{73.8}{2.5}{}} & \tabresult{66.6}{6.5}{} & \textbf{\tabresult{79.4}{9.8}{}} \\
\bottomrule
\end{tabular*}
\end{table*}

\begin{table*}[!t]
\centering
\footnotesize
\caption{Results (\%) of MCDA and its three variants in ASD vs. HC classification. Best results are shown in bold. Here, * denotes a statistically significant (\textit{p}\textless 0.05) difference between MCDA and a variant in terms of accuracy.}
\label{tab:ablation study of ASD}

\setlength{\tabcolsep}{1.25pt}

\begin{tabular*}{\textwidth}{@{\extracolsep{\fill}}lcccccccccccc@{}}
\toprule
\multirow{2}{*}{Method}
& \multicolumn{4}{c}{Site NYU}
& \multicolumn{4}{c}{Site UM}
& \multicolumn{4}{c}{Site USM} \\
\cmidrule(lr){2-5} \cmidrule(lr){6-9} \cmidrule(lr){10-13}
& ACC & AUC & PRE & REC
& ACC & AUC & PRE & REC
& ACC & AUC & PRE & REC \\
\midrule
MCDA-I &
\tabresult{76.0}{7.1}{} & \tabresult{77.5}{8.3}{} & \tabresult{70.2}{11.9}{} & \tabresult{58.0}{10.4}{} &
\tabresult{70.9}{7.6}{} & \tabresult{73.0}{8.3}{} & \tabresult{78.0}{18.0}{} & \tabresult{59.4}{26.2}{} &
\tabresult{74.5}{2.6}{} & \tabresult{76.1}{4.1}{} & \tabresult{76.0}{6.9}{} & \tabresult{74.7}{6.8}{} \\
MCDA-II &
\tabresult{73.6}{6.8}{*} & \tabresult{76.0}{7.4}{} & \textbf{\tabresult{81.7}{7.6}{}} & \tabresult{57.6}{14.2}{} &
\tabresult{72.7}{4.5}{} & \tabresult{75.7}{7.2}{} & \tabresult{81.9}{6.4}{} & \tabresult{61.3}{16.3}{} &
\tabresult{70.2}{5.3}{*} & \tabresult{70.6}{7.9}{} & \tabresult{64.9}{9.9}{} & \tabresult{78.4}{9.1}{} \\
MCDA-III &
\tabresult{69.0}{9.0}{*} & \tabresult{68.0}{6.5}{} & \tabresult{77.0}{16.3}{} & \tabresult{61.5}{14.9}{} &
\tabresult{65.4}{3.3}{*} & \tabresult{65.4}{1.7}{} & \tabresult{65.7}{7.3}{} & \textbf{\tabresult{69.8}{2.1}{}} &
\tabresult{67.4}{5.2}{*} & \tabresult{66.4}{8.0}{} & \tabresult{66.2}{6.3}{} & \tabresult{75.3}{10.6}{} \\
\midrule
MCDA &
\textbf{\tabresult{81.1}{2.1}{}} & \textbf{\tabresult{82.9}{2.4}{}} & \tabresult{76.1}{9.9}{} & \textbf{\tabresult{68.1}{8.0}{}} &
\textbf{\tabresult{75.0}{2.6}{}} & \textbf{\tabresult{78.4}{7.2}{}} & \textbf{\tabresult{87.4}{3.6}{}} & \tabresult{61.2}{9.6}{} &
\textbf{\tabresult{80.2}{7.1}{}} & \textbf{\tabresult{82.1}{5.8}{}} & \textbf{\tabresult{77.9}{8.9}{}} & \textbf{\tabresult{86.2}{8.4}{}} \\
\bottomrule
\end{tabular*}
\end{table*}

To assess the contribution of key components in our MCDA framework, we compare it with three ablated variants, namely \textbf{MCDA-I}, \textbf{MCDA-II}, and \textbf{MCDA-III}. Specifically, MCDA-I removes the DAU module and directly uses the generated self-aware and cross-sample augmentation views without history-guided refinement. 
In MCDA-II, to evaluate the impact of the contrastive loss \(\mathcal{L}_{\text{SA,CA}}\), we set the weighting parameter \(\beta\) in Eq.~\eqref{eq:TMC loss} to 0. 
Similarly, MCDA-III sets $\lambda=0$ in Eq.~\eqref{eq:loss}, while retaining both augmentation branches, the DAU module, and the classification loss computed from the averaged prediction probabilities of the three views.
The same three-view probability averaging is used during inference, isolating the contribution of the contrastive objective.
These variants are trained on the two datasets with the same experiment setting, with results reported in Table~\ref{tab:ablation study of MDD} for MDD vs. HC classification and Table~\ref{tab:ablation study of ASD} for ASD vs. HC classification. 

The complete MCDA model achieves the highest mean ACC and AUC in all six within-site evaluations. Its improvement over MCDA-I supports the contribution of the DAU module. The higher mean ACC and AUC compared with MCDA-II further support the benefit of including the \(\mathcal{L}_{\text{SA,CA}}\) diversity constraint. Meanwhile, the comparison with the classification-only baseline MCDA-III indicates the additional value of multi-view contrastive regularization. Overall, these ablations support the contributions of the evaluated components to within-site classification performance.

\begin{figure*}[!t]
    \centering\includegraphics[width=\textwidth]{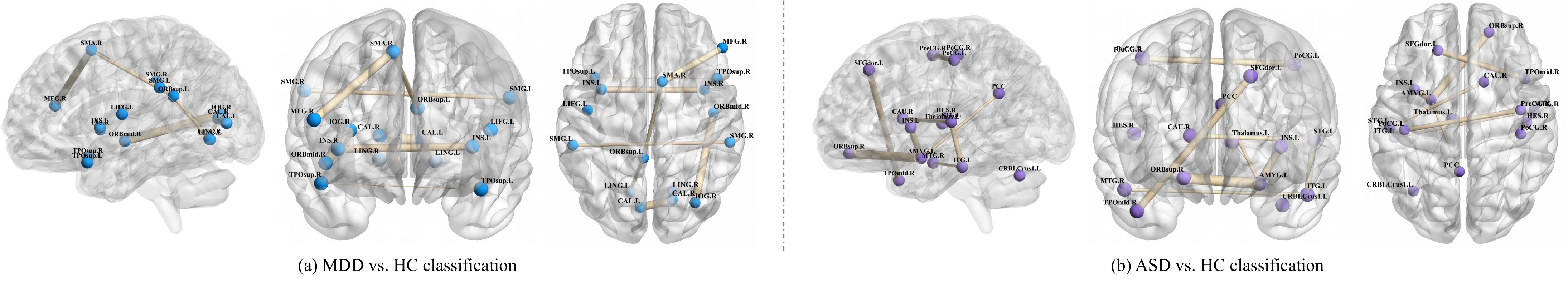}
    \vspace{-1.5em}
    \caption{Visualization of the top 10 ROI-pair associations identified by MCDA in two classification tasks: (a) MDD vs. HC classification (Site 20) on REST-MDD, and (b) ASD vs. HC classification (Site NYU) on ABIDE.}
    \label{fig:Visualization of top 10 functional connectivity}
    \vspace{-1em}
\end{figure*}

\section{Discussion}\label{sec:discussion}

\subsection{Exploratory Analysis of Brain Region-Level Associations}

We analyze brain region-level associations in the self-aware and cross-sample augmentation features of held-out test subjects from Site 20 of REST-MDD and Site NYU of ABIDE separately. Features are extracted using frozen models trained and selected without the corresponding test subjects. For each subject and view, ROI temporal-by-channel features are centered over time, and linear centered kernel alignment (CKA)~\cite{kornblith2019similarity} is used to construct a 116$\times$116 association matrix. For each ROI pair, the two view-specific scores form a two-dimensional response in a multivariate linear model adjusted for prespecified demographic and head-motion covariates. Diagnostic-group effects are evaluated using Pillai's trace and Freedman--Lane residual permutations~\cite{winkler2014permutation}. Covariate-only model residuals are permuted jointly across subjects for both views and added back to the fitted values before refitting, using identical permutations for all ROI pairs. Joint $p$-values are corrected across the 6,670 unique ROI pairs within each site using the Benjamini--Yekutieli procedure~\cite{benjamini2001control}. Among ROI pairs with BY-adjusted $p$-values below 0.05, up to ten pairs with the smallest joint $p$-values are
visualized using BrainNet Viewer~\cite{xia2013brainnet},
with line thickness representing Pillai's trace. The MDD--HC and ASD--HC patterns are shown in Fig.~\ref{fig:Visualization of top 10 functional connectivity}(a) and (b), respectively.

As shown in Fig.~\ref{fig:Visualization of top 10 functional connectivity}(a), the visualized connections for the MDD--HC comparison involve the bilateral insula (INS.L and INS.R), right middle frontal gyrus (MFG.R), and right supplementary motor area (SMA.R). Previous studies have associated insula network connectivity with depressive symptoms in MDD~\cite{he2022insula} and reported altered structural--functional coupling of the right SMA~\cite{wei2023disrupted}.
For the ASD--HC comparison, the connections visualized in Fig.~\ref{fig:Visualization of top 10 functional connectivity}(b) involve the left amygdala (AMYG.L), right superior frontal gyrus, orbital part (ORBsup.R), and right precentral gyrus (PreCG.R). Previous rs-fMRI research has reported atypical amygdala connectivity in children and adolescents with ASD~\cite{fishman2018atypical}.
Overall, these observations provide exploratory anatomical context for the visualized brain region associations.

\subsection{Visualization of Dynamic Graph Augmentation}
\begin{figure*}[!t]
    \centering \includegraphics[width=\textwidth]{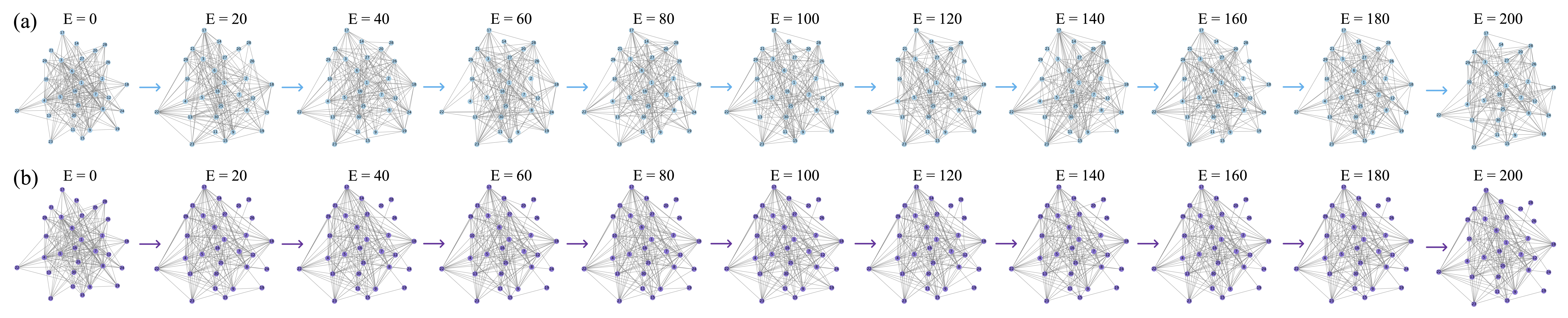}
    \vspace{-2em}
    \caption{Visualization of dynamic augmented brain networks by DAU module at every 20 epochs in (a) self-aware and (b) cross-sample augmentation.}
    \label{fig:visualization of dynamic graph augmentation}
\end{figure*}

To elucidate the process of two dynamic graph topology augmentations, we visualize the generated brain network augmentation. 
Specifically, taking a dynamic self-aware augmented brain network of the first 30 ROIs from a subject in Site 20 of the REST-MDD as an example, along with the original graph at E=0, are visualized at a step interval of 20, with results presented in Fig.~\ref{fig:visualization of dynamic graph augmentation}(a). 
Similarly, we show 11 graphs generated in a dynamic cross-sample augmented brain network with the same subject in Fig.~\ref{fig:visualization of dynamic graph augmentation}(b). 

As illustrated in Fig.~\ref{fig:visualization of dynamic graph augmentation}, the two augmented brain networks maintain strong structural consistency during the dynamic augmentation process. This demonstrates that the backward propagation in DAU module effectively constrains and adjusts the consistency of semantic representations, ensuring meaningful topological transformations while preserving critical functional connectivity patterns. 
Meanwhile, from a topological structure perspective, the self-aware augmented networks maintain high similarity with the original networks. In contrast, the cross-sample augmented networks demonstrate more substantial modifications, particularly in the connections between ROIs located in the upper-right and lower-left regions. 

\subsection{Influence of Dynamic Augmentation Update Structure}

\begin{figure}[!t]
    \centering\includegraphics[width=0.85\columnwidth]{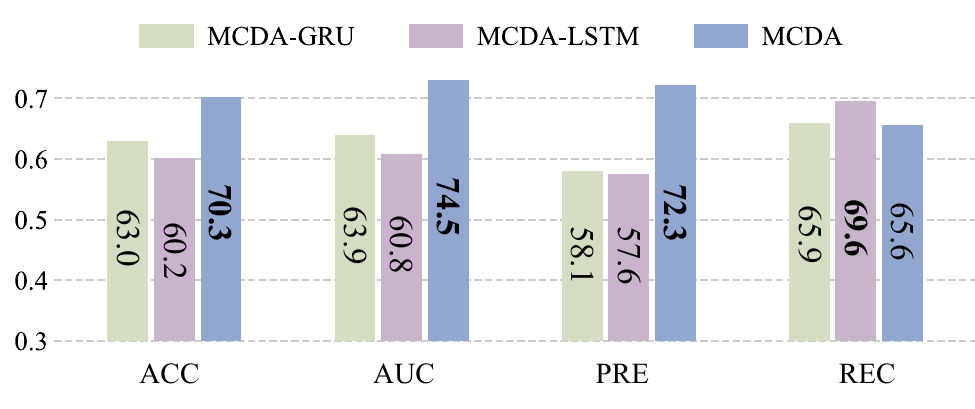}
    \vspace{-1em}
    \caption{Results of MCDA-GRU, MCDA-LSTM and MCDA on Site 20 from the REST-MDD dataset.}
    \label{fig:Influence of Dynamic Update Augmentation Structure}
\end{figure}

To investigate the effect of different dynamic update structures on the performance of our MCDA framework, the proposed DAU module is replaced with GRU~\cite{yang2024early} and LSTM~\cite{supakar2022deep}, referred to as \textbf{MCDA-GRU} and \textbf{MCDA-LSTM}, respectively. 
Specifically, the MCDA-GRU employs a standard 3-layer GRU architecture with forward propagation only, while the MCDA-LSTM utilizes a 3-layer LSTM structure, both maintaining consistent configurations with our DAU module to ensure fair comparison. 
We conduct experiments on Site 20 of REST-MDD and report results in Fig.~\ref{fig:Influence of Dynamic Update Augmentation Structure}. As can be seen from this figure, our MCDA framework consistently outperforms MCDA-GRU and MCDA-LSTM across most evaluation metrics, highlighting the effectiveness of the bidirectional dynamic update mechanism for promoting the propagation of discriminative information. 
The ability to maintain semantic consistency in the topology augmentation during continuous updates plays a key role.

\vspace{-1em}
\begin{table}[!t]
\setlength{\tabcolsep}{3pt}
\footnotesize
\centering
\caption{Results (\%) of different augmentation encoders on Site 20 from the REST-MDD dataset.}
\label{tab:influence of augmentation encoder on site20}

\begin{tabularx}{\linewidth}{lZZZZ}
\toprule
Method & ACC & AUC & PRE & REC \\
\midrule
MCDA-None &
59.8\textsubscript{\text{\tiny ±4.3}} & 60.0\textsubscript{\text{\tiny ±6.8}} & 52.8\textsubscript{\text{\tiny ±8.3}} & 53.7\textsubscript{\text{\tiny ±7.3}} \\
MCDA-Ra-GConv &
64.2\textsubscript{\text{\tiny ±3.7}} & 66.2\textsubscript{\text{\tiny ±4.2}} & 57.1\textsubscript{\text{\tiny ±6.2}} & \textbf{67.9\textsubscript{\text{\tiny ±5.9}}} \\
MCDA-2DCNN &
62.6\textsubscript{\text{\tiny ±3.0}} & 65.0\textsubscript{\text{\tiny ±4.6}} & 58.3\textsubscript{\text{\tiny ±4.4}} & 66.0\textsubscript{\text{\tiny ±3.9}} \\
\midrule
MCDA &
\textbf{70.3\textsubscript{\text{\tiny ±2.9}}} & \textbf{74.5\textsubscript{\text{\tiny ±2.8}}} & \textbf{72.3\textsubscript{\text{\tiny ±4.0}}} & 65.6\textsubscript{\text{\tiny ±8.4}} \\
\bottomrule
\end{tabularx}
\vspace{-1.5em}
\end{table}

\subsection{Influence of Augmentation Encoder}

To evaluate the contribution of the ST-GC encoder in capturing spatial and temporal brain connectivity, we conduct ablation studies by replacing it with alternative modules. 
Specifically, we substitute the ST-GC encoder with the Ra-GConv module~\cite{li2021braingnn} and a conventional 2D convolutional module~\cite{hu2019multichannel}, resulting in three variants: \textbf{MCDA-None} (ST-GC encoder removed), \textbf{MCDA-Ra-GConv} (replaced with Ra-GConv), and \textbf{MCDA-2DCNN} (replaced with 2D CNN). 
Ra-GConv leverage the topological and functional information of fMRI to learn different embedding weights in graph convolutional kernels conditioned on the ROI (geometrically distributed information of the brain), while 2D CNN operations process rs-fMRI data as regular grid-like inputs with sliding a fixed-size convolution kernel to extract localized patterns on the graph. 
Consistent with the ST-GC encoder 3-layer structure, we implement both Ra-GConv and 2D-CNN variants with three layers each. 
The results of MCDA and three variants on Site 20 of REST-MDD are reported in Table~\ref{tab:influence of augmentation encoder on site20}. 
This table suggests that the MCDA-Ra-GConv and MCDA-2DCNN variants show decent yet limited effectiveness; however, MCDA-None yields significantly inferior cross-validation results, primarily due to its inability to effectively extract and utilize information from the augmented brain networks. 
In contrast, our proposed ST-GC encoder outperforms all three variants by a clear margin. These results suggest that fully leveraging both temporal and spatial attributes of brain networks enables the extraction of more informative and diagnostically relevant features, thereby enhancing the performance of downstream classification tasks.

\subsection{Parameter Analysis}
Site 20 of REST-MDD was used as the development site. Hyperparameters were selected using validation AUC from the five-fold partitions of this site. The selected configuration was subsequently fixed for all site-wise experiments. The reported results reveal performance on the held-out test folds for the evaluated parameter settings.

\subsubsection{Influence of Trade-off TMC Loss Parameters}

\begin{figure}[!t]
    \centering    
    \includegraphics[width=\columnwidth]{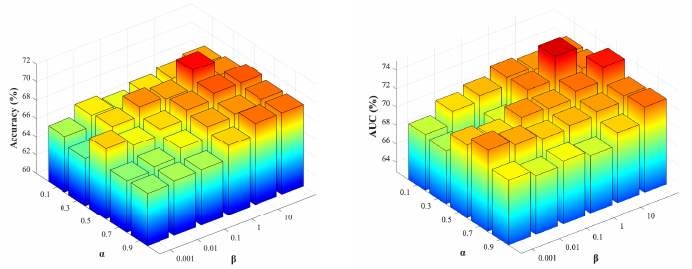}
    \vspace{-1.5em}
    \caption{Accuracy and AUC of MCDA with different Topology-guided Multi-view Contrastive Loss parameters $\alpha$ and $\beta$ on Site 20 from the REST-MDD dataset.}
    \label{fig:Influence of alpha and beta}
    \vspace{-2em}
\end{figure}

\begin{figure}[!t]
    \centering
    \includegraphics[width=\columnwidth]{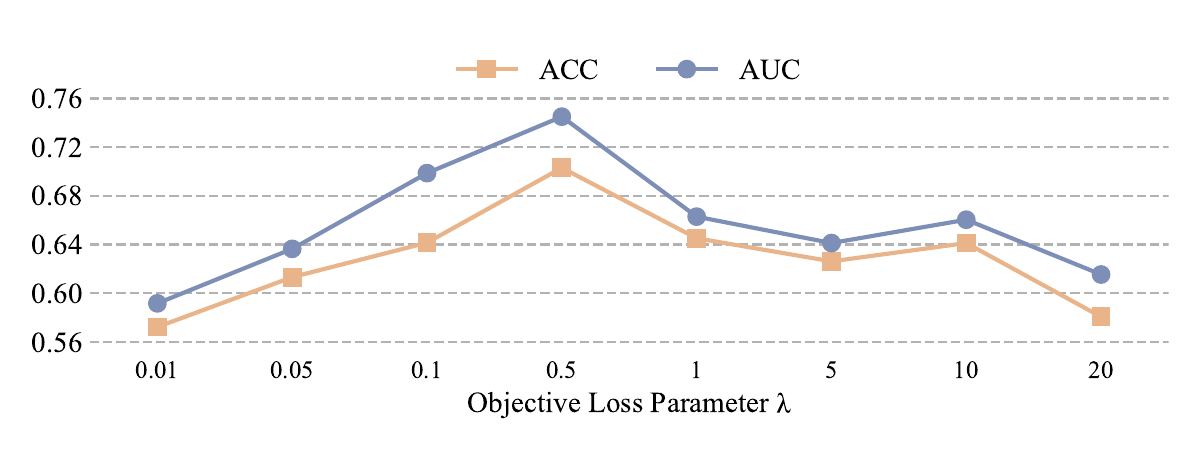}
    \vspace{-2.5em}
    \caption{Results of MCDA with different loss parameter $\lambda$ on Site 20 from the REST-MDD dataset.}
    \label{fig:Influence of Loss Param}
    \vspace{-1em}
\end{figure}

To examine the effects of trade-off parameters in topology-guided multi-view contrastive loss $\mathcal{L}_{\text{TMC}}$ of Eq.~\eqref{eq:TMC loss}, we adjust the values of $\alpha$ and $\beta$. The parameter $\alpha$, which regulates the contribution of self-aware and cross-sample augmentation, is varied from 0 to 1. 
Meanwhile, $\beta$, reflecting the impact of differences between augmented brain networks, is tested at 0.001, 0.01, 0.1, 1, and 10. Fig.~\ref{fig:Influence of alpha and beta} presents a subset of the accuracy and AUC results, which indicates that MCDA achieves optimal performance when $\alpha$ = 0.3, $\beta$ = 1. 
Furthermore, we investigate how the TMC loss \(\mathcal{L}_{\text{TMC}}\), scaled by the trade-off parameter \(\lambda\), affects the classification performance. The corresponding results on Site 20 are represented in Fig.~\ref{fig:Influence of Loss Param}, and MCDA ultimately achieves its optimum at $\lambda $ = 0.5. Using large values for $\lambda$, our framework cannot achieve good results. This implies that excessive weighting of the TMC loss causes the framework to overly focus on the relative distances between multiple augmented views, leading to a less compact feature embedding space and consequently impairing classification performance.

\subsubsection{Influence of History Windows Size}\label{sec:influece of historical windows size}

As shown in Table~\ref{tab:history_window-site-site20}, increasing $K$ from 1 to 5 improves ACC from 60.23\% to 70.28\% and AUC from 56.36\% to 74.53\%, with $K$=5 also achieving the highest PRE of 72.29\%. A moderate descriptor window may provide useful cross-epoch context for BiGRU-based gate estimation, whereas $K$=1 supplies only the current descriptor. Extending the window beyond five epochs does not further improve these three metrics, possibly because older descriptors reflect earlier training states and may be less compatible with the current augmentation. Although $K$=9 achieves a slightly higher REC than $K$=5, $K$=5 provides better ACC, AUC, and PRE.

\subsubsection{Influence of Batch Size}\label{sec:influece of batch size}
\begin{table}[!t]
\setlength{\tabcolsep}{3pt}
\footnotesize
\centering
\caption{Results (\%) of MCDA with different history window lengths $K$ on Site 20 from the REST-MDD dataset.}
\label{tab:history_window-site-site20}

\begin{tabularx}{\linewidth}{>{\raggedright\arraybackslash}p{0.14\linewidth}ZZZZ}
\toprule
$K$ & ACC & AUC & PRE & REC \\
\midrule
1 &
60.2\textsubscript{\text{\tiny ±3.1}} & 56.4\textsubscript{\text{\tiny ±3.7}} & 56.0\textsubscript{\text{\tiny ±5.7}} & 65.5\textsubscript{\text{\tiny ±2.4}} \\
3 &
64.4\textsubscript{\text{\tiny ±0.8}} & 59.2\textsubscript{\text{\tiny ±3.5}} & 61.0\textsubscript{\text{\tiny ±3.5}} & 61.4\textsubscript{\text{\tiny ±6.7}} \\
5 &
\textbf{70.3\textsubscript{\text{\tiny ±2.9}}} & \textbf{74.5\textsubscript{\text{\tiny ±2.8}}} & \textbf{72.3\textsubscript{\text{\tiny ±4.0}}} & 65.6\textsubscript{\text{\tiny ±8.4}} \\
7 &
67.0\textsubscript{\text{\tiny ±2.9}} & 66.6\textsubscript{\text{\tiny ±9.6}} & 67.7\textsubscript{\text{\tiny ±6.9}} & 62.1\textsubscript{\text{\tiny ±6.5}} \\
9 &
69.0\textsubscript{\text{\tiny ±3.5}} & 68.0\textsubscript{\text{\tiny ±8.5}} & 66.9\textsubscript{\text{\tiny ±6.5}} & \textbf{65.8\textsubscript{\text{\tiny ±5.1}}} \\
\bottomrule
\end{tabularx}
\vspace{-1em}
\end{table}

\begin{table}[!t]
\setlength{\tabcolsep}{3pt}
\footnotesize
\centering
\caption{Results (\%) of MCDA with different batch sizes on Site 20 from the REST-MDD dataset.}
\label{tab:batch-site-site20}

\begin{tabularx}{\linewidth}{>{\raggedright\arraybackslash}p{0.16\linewidth}ZZZZ}
\toprule
Batch Size & ACC & AUC & PRE & REC \\
\midrule
30 &
59.1\textsubscript{\text{\tiny ±6.1}} & 56.2\textsubscript{\text{\tiny ±4.6}} & 53.2\textsubscript{\text{\tiny ±15.4}} & 44.5\textsubscript{\text{\tiny ±14.7}} \\
35 &
61.6\textsubscript{\text{\tiny ±6.4}} & 60.4\textsubscript{\text{\tiny ±5.3}} & 57.3\textsubscript{\text{\tiny ±4.8}} & 54.3\textsubscript{\text{\tiny ±8.8}} \\
40 &
58.9\textsubscript{\text{\tiny ±6.2}} & 61.4\textsubscript{\text{\tiny ±10.1}} & 55.3\textsubscript{\text{\tiny ±9.3}} & 50.4\textsubscript{\text{\tiny ±7.8}} \\
45 &
\textbf{70.3\textsubscript{\text{\tiny ±2.9}}} & \textbf{74.5\textsubscript{\text{\tiny ±2.8}}} & \textbf{72.3\textsubscript{\text{\tiny ±4.0}}} & \textbf{65.6\textsubscript{\text{\tiny ±8.4}}} \\
50 &
64.2\textsubscript{\text{\tiny ±4.0}} & 67.1\textsubscript{\text{\tiny ±7.5}} & 52.7\textsubscript{\text{\tiny ±7.5}} & 65.1\textsubscript{\text{\tiny ±8.2}} \\
\bottomrule
\end{tabularx}
\vspace{-1.5em}
\end{table}

We further study the impact of different batch sizes. Based on data from Site 20 of REST-MDD, we set variations of 30 to 50 with a interval of 5 to assess the impact of subject variability across different group sizes on classification tasks. Specifically, the results on Site 20 are presented Table~\ref{tab:batch-site-site20}. 
The optimal performance across the four metrics is achieved when the batch size is set to 45. Using an excessively large or small batch size limits the ability of dynamic topology augmentation to capture inter-subject variability, which in turn reduces its effectiveness in identifying discriminative features.

\subsection{Limitation and Future Work}\label{sec: limitation and furture work}

Several limitation of current work need to be considered. 
\emph{On one hand}, the brain network construction relies solely on rs-fMRI. In the future, we will integrate multi-modal data (e.g., diffusion tensor imaging~\cite{zhang2021deep}, demographic factors~\cite{hernandez2024brain} and clinical scores~\cite{mishra2021review}) to explore complementary relationships between imaging and none-imaging patterns. 
\emph{On the other hand}, our application scenario is limited to data collected from a single site, overlooking the potential heterogeneity caused by differences in acquisition machine parameters across multiple sites in real-world settings. 
A possible solution is to employ domain adaption to reduce cross-domain distribution difference~\cite{guan2021domain}, which will be one of our future work. 

\section{Conclusion}\label{sec:conclusion}

This paper introduces a topology-guided multi-view graph contrastive learning framework (MCDA) with dynamic topology augmentation for brain disorder diagnosis and analysis. In our design, two complementary brain networks are generated through a dynamic augmentation update (DAU) module, incorporating both self-aware and cross-sample strategies to capture subject-specific traits and functional heterogeneity across individuals. In addition, we develop a topology-aware multi-view contrastive learning paradigm that simultaneously models the interrelationships between the original graph and its augmented counterparts, as well as between the augmented views themselves. We further characterize the optimization directions of the contrastive constraints through similarity-score gradient analysis and establish a lower bound for the contrastive objective.
Experiments on two public rs-fMRI datasets demonstrate improved classification performance compared with several state-of-the-art methods.

% \section*{Declaration of Competing Interest}
% The authors declare that they have no known competing financial interests or personal relationships that could have appeared to influence the work reported in this paper.

% \section*{Data Availability}
% The code has been made publicly available at: \url{https://github.com/goodcodingagoodboy/MCDA}. All the leveraged datasets and methods are publicly available online.

\appendix[Optimization-Direction Analysis of the Multi-View Contrastive Objective]
\label{app:optimization_direction}

This appendix examines the optimization directions and boundedness of the proposed multi-view contrastive objective.
The analysis is conducted with respect to the similarity scores and provides a geometric interpretation under unit-norm normalization.
\vspace{-1em}
\subsection{Similarity scores and loss formulation}

The three constraints operate on the counterparts, respectively. Their similarity scores are defined as follows:
\begin{equation}
\begin{aligned}
s_{ij}^{\text{SA}}&=\operatorname{sim}\left(\mathbf{G}_{i}^{\text{SA}},\mathbf{G}_{j}\right),\\
s_{ij}^{\text{CA}}&=\operatorname{sim}\left(\mathbf{G}_{i}^{\text{CA}},\mathbf{G}_{j}\right),\\
s_{ij}^{\text{SA,CA}}&=\operatorname{sim}\left(\mathbf{G}_{i}^{\text{SA}},\mathbf{G}_{j}^{\text{CA}}\right),
\end{aligned}
\label{eq:app_similarity_scores}
\end{equation}

For each constraint indexed by $r\in\{\text{SA},\text{CA},(\text{SA,CA})\}$, the similarities are normalized across the mini-batch.
The resulting matching score is defined as follows:
\begin{equation}
q_{ij}^{r}
=
\frac{\exp(s_{ij}^{r}/\tau)}
{\sum_{k=1}^{B}\exp(s_{ik}^{r}/\tau)},
\label{eq:app_matching_scores}
\end{equation}

The quantity $q_{ij}^{r}$ represents the relative matching score within the mini-batch and is distinct from the classification probability $p_i$.
Throughout this analysis, we assume $B\geq2$, $\tau>0$, and finite similarity scores.
Under these conditions, every matching score lies strictly between zero and one.

The three losses differ in the sign assigned to the matched-pair log-ratio. They can therefore be expressed in the following unified form:
\begin{equation}
\begin{aligned}
\mathcal{L}_{r}&=\frac{\eta_r}{B}\sum_{i=1}^{B}\log q_{ii}^{r},\quad
\eta_{\text{SA}}=-1,\quad\eta_{\text{CA}}=\eta_{\text{SA,CA}}=1.
\end{aligned}
\label{eq:app_unified_loss}
\end{equation}

The negative sign specifies the consistency constraint for the self-aware with original pair.
The positive signs specify the diversity constraints for the cross-sample with original and self-aware with cross-sample pairs.
\vspace{-1em}
\subsection{Optimization directions}

The logarithm of the matched score admits the following expansion:
\begin{equation}
\log q_{ii}^{r}
=
\frac{s_{ii}^{r}}{\tau}
-
\log\left[
\sum_{k=1}^{B}\exp(s_{ik}^{r}/\tau)
\right],
\label{eq:app_log_matching_score}
\end{equation}

Differentiating the unified loss with respect to an individual similarity score yields the following expression:
\begin{equation}
\frac{\partial\mathcal{L}_{r}}
{\partial s_{ij}^{r}}
=
\frac{\eta_r}{B\tau}
\left(\delta_{ij}-q_{ij}^{r}\right),
\label{eq:app_general_gradient}
\end{equation}

Here, $\delta_{ij}$ is the Kronecker delta. It equals one for matched pairs and zero for non-matched pairs. For matched pairs, setting $j=i$ gives the following derivatives:
\begin{equation}
\begin{aligned}
\frac{\partial\mathcal{L}_{\text{SA}}}{\partial s_{ii}^{\text{SA}}}&=-\frac{1-q_{ii}^{\text{SA}}}{B\tau}<0,\\
\frac{\partial\mathcal{L}_{\text{CA}}}{\partial s_{ii}^{\text{CA}}}&=\frac{1-q_{ii}^{\text{CA}}}{B\tau}>0,\\
\frac{\partial\mathcal{L}_{\text{SA,CA}}}{\partial s_{ii}^{\text{SA,CA}}}&=\frac{1-q_{ii}^{\text{SA,CA}}}{B\tau}>0,
\end{aligned}
\label{eq:app_matched_gradients}
\end{equation}

The strict signs follow from the matching scores lying between zero and one.
When the similarity scores are treated as independent coordinates, the negative-gradient directions favor a higher matched self-aware with original score and lower matched cross-sample with original and self-aware with cross-sample scores.

For non-matched pairs, setting $j\neq i$ gives the following derivatives:
\begin{equation}
\begin{aligned}
\frac{\partial\mathcal{L}_{\text{SA}}}{\partial s_{ij}^{\text{SA}}}&=\frac{q_{ij}^{\text{SA}}}{B\tau}>0,\\
\frac{\partial\mathcal{L}_{\text{CA}}}{\partial s_{ij}^{\text{CA}}}&=-\frac{q_{ij}^{\text{CA}}}{B\tau}<0,\\
\frac{\partial\mathcal{L}_{\text{SA,CA}}}{\partial s_{ij}^{\text{SA,CA}}}&=-\frac{q_{ij}^{\text{SA,CA}}}{B\tau}<0.
\end{aligned}
\label{eq:app_nonmatched_gradients}
\end{equation}
The self-aware constraint favors lower non-matched scores, whereas the two diversity constraints favor higher non-matched scores.
Together with the matched-pair derivatives, these results show that the losses regulate matched scores relative to other pairings within the mini-batch.
In particular, the diversity constraints do not impose uniform repulsion between all views.

The combined objective weights the three constraints by $\alpha$, $1-\alpha$, and $\beta$, respectively.
For $\alpha\in[0,1]$ and $\beta\geq0$, positive weights preserve the corresponding derivative signs in similarity-score space.
A zero weight removes the associated constraint.
\vspace{-1em}
\subsection{Geometric interpretation}

Cosine similarity provides a distance interpretation when the norms of the compared views are at least $\varepsilon>0$.
Under this condition, the numerical-stability operation does not alter the normalization, and cosine similarity equals the inner product of unit-normalized representations.
For the self-aware with original pair, the normalized representations are defined as follows:
\begin{equation}
\begin{aligned}
\widehat{\mathbf{G}}_{i}&=\frac{\mathbf{G}_{i}}{\|\mathbf{G}_{i}\|_{T,N}},\quad
\widehat{\mathbf{G}}_{i}^{\text{SA}}=\frac{\mathbf{G}_{i}^{\text{SA}}}{\|\mathbf{G}_{i}^{\text{SA}}\|_{T,N}},
\end{aligned}
\label{eq:app_normalized_views}
\end{equation}

Expanding their squared distance gives the following identity:
\begin{equation}
\left\|
\widehat{\mathbf{G}}_{i}^{\text{SA}}
-
\widehat{\mathbf{G}}_{i}
\right\|_{T,N}^{2}
=
2-2s_{ii}^{\text{SA}}.
\label{eq:app_cosine_distance_relation}
\end{equation}
The same identity applies to the two remaining pairs under the corresponding normalization conditions.
A higher cosine similarity therefore corresponds to a smaller squared distance between unit-normalized representations, and a lower similarity corresponds to a larger distance.
This identity supplies a geometric interpretation of the score-level optimization directions.
It does not establish corresponding changes in distances between unnormalized representations, graph topology, or diagnostic semantics.
\vspace{-1em}
\subsection{Boundedness of the objective}

The cosine similarity scores lie between minus one and one.
Factoring the matched exponential term out of the denominator rewrites the matching score as follows:
\begin{equation}
q_{ii}^{r}
=
\left[
1+
\sum_{\substack{k=1\\k\neq i}}^{B}
\exp\left(
\frac{s_{ik}^{r}-s_{ii}^{r}}{\tau}
\right)
\right]^{-1},
\label{eq:app_matching_score_rewritten}
\end{equation}
Each difference between a non-matched and matched score is at most two.
Applying this bound to the denominator yields the following inequality:
\begin{equation}
\frac{1}{1+(B-1)\exp(2/\tau)}
\leq q_{ii}^{r}<1,
\label{eq:app_matching_score_bound}
\end{equation}
Taking logarithms and applying the negative sign of the SA constraint gives the following consistency-loss bound:
\begin{equation}
0\leq\mathcal{L}_{\text{SA}}
\leq
\log\left[1+(B-1)\exp(2/\tau)\right],
\label{eq:app_consistency_loss_bound}
\end{equation}
The positive signs of the CA and SA--CA constraints instead give the following diversity-loss bounds:
\begin{equation}
\begin{gathered}
-\log\left[1+(B-1)\exp(2/\tau)\right]\leq\mathcal{L}_{r}\leq0,
% r\in\{\text{CA},(\text{SA,CA})\},
\end{gathered}
\label{eq:app_diversity_loss_bound}
\end{equation}
The consistency loss is non-negative.
Combining the two diversity-loss lower bounds with their non-negative weights therefore gives a lower bound for the complete contrastive objective:
\begin{equation}
\mathcal{L}_{\text{TMC}}
\geq
-(1-\alpha+\beta)
\log\left[1+(B-1)\exp(2/\tau)\right].
\label{eq:app_combined_loss_bound}
\end{equation}

Thus, although the diversity terms can take negative values, the contrastive objective cannot decrease without bound for fixed $B$, $\tau$, $\alpha$, and $\beta$.
This property does not, by itself, guarantee optimization stability or convergence.

\bibliographystyle{IEEEtran}

\bibliography{reference}

\end{document}